\documentclass[nohyperref]{article}

\usepackage{microtype}
\usepackage{graphicx}
\usepackage{subfigure}
\usepackage{booktabs} 

\newcommand{\ack}[1]{SPED}

\usepackage{bbold}

\usepackage{cancel}

\usepackage[accepted]{icml2022_TAGML}

\usepackage{amsmath}
\usepackage{amssymb}
\usepackage{mathtools}
\usepackage{amsthm}

\usepackage[capitalize,noabbrev]{cleveref}

\theoremstyle{plain}

\theoremstyle{definition}

\theoremstyle{remark}

\usepackage[textsize=tiny]{todonotes}

\icmltitlerunning{Stochastic Parallelizable Eigengap Dilation for Large Graph Clustering}

\begin{document}

\twocolumn[
\icmltitle{Stochastic Parallelizable Eigengap Dilation for Large Graph Clustering}



\icmlsetsymbol{equal}{*}

\begin{icmlauthorlist}
\icmlauthor{Elise van~der~Pol}{univ}
\icmlauthor{Ian Gemp}{dm}
\icmlauthor{Yoram Bachrach}{dm}
\icmlauthor{Richard Everett}{dm}
\end{icmlauthorlist}

\icmlaffiliation{univ}{University of Amsterdam, Amsterdam, Netherlands. Work completed while at DeepMind.}
\icmlaffiliation{dm}{DeepMind, London, UK}

\icmlcorrespondingauthor{Elise van~der~Pol}{elisevanderpol@gmail.com}
\icmlcorrespondingauthor{Ian Gemp}{imgemp@deepmind.com}

\icmlkeywords{EigenGame, Graph Clustering}

\vskip 0.3in
]



\printAffiliationsAndNotice{}  

\begin{abstract}
Large graphs commonly appear in social networks, knowledge graphs, recommender systems, life sciences, and decision making problems. Summarizing large graphs by their high level properties is helpful in solving problems in these settings. In spectral clustering, we aim to identify clusters of nodes where most edges fall within clusters and only few edges fall between clusters. This task is important for many downstream applications and exploratory analysis. A core step of spectral clustering is performing an eigendecomposition of the corresponding graph Laplacian matrix (or equivalently, a singular value decomposition, SVD, of the incidence matrix).
The convergence of iterative singular value decomposition approaches depends on the eigengaps of the spectrum of the given matrix, i.e., the difference between consecutive eigenvalues. For a graph Laplacian corresponding to a well-clustered graph, the eigenvalues will be non-negative but very small (much less than $1$) slowing convergence.
This paper introduces a parallelizable approach to dilating the spectrum in order to accelerate SVD solvers and in turn, spectral clustering. This is accomplished via polynomial approximations to matrix operations that favorably transform the spectrum of a matrix without changing its eigenvectors. Experiments demonstrate that this approach significantly accelerates convergence, and we explain how this transformation can be parallelized and stochastically approximated to scale with available compute.
\end{abstract}

\section{Introduction}\label{sec:intro}
Large graphs commonly appear in social networks, knowledge graphs, recommender systems, life sciences, and decision making problems. In spectral clustering, we aim to identify node clusters so that most edges fall within clusters and only few edges fall between clusters. This task is important for many downstream applications such as community detection in sociology or biology~\citep{fortunato2010community}, image segmentation~\citep{coleman1979image}, generating or refining labels~\citep{song2015spectral}, exploratory data analysis~\citep{kumar2020visual}, and more abstractly, approximating solutions to combinatorial graph problems like min-cut~\citep{chung1997spectral}.


A key intermediate step in spectral clustering~\citep{von2007tutorial} performs an eigendecomposition of the corresponding graph Laplacian matrix. Specifically, the eigenvectors associated with the bottom-$k$ eigenvalues of the graph Laplacian define $k$-dimensional embeddings for each node in the graph providing a ``soft'' clustering. This resulting embedding well-separates nodes that belong in separate clusters, making a final ``hard'' clustering step, e.g., with $k$-means, relatively trivial.

For large graphs, computing a full eigendecomposition can be computationally costly, as doing an eigendecomposition of the graph Laplacian has complexity $\mathcal{O}(\vert \mathcal{E} \vert \vert \mathcal{V} \vert^2)$ where $\vert \mathcal{V} \vert$ is the number of nodes and $\vert \mathcal{E} \vert$ the number of edges~\citep{allen2017first}. In practice, we are only interested in a subset of $k$ eigenvectors. The convergence rate of many iterative solvers depends on the normalized eigengap,  i.e. the difference between consecutive eigenvalues relative to the spectral radius~\citep{balcan2016improved,gemp2020eigengame}. For the bottom-$k$ eigenvalues of a large graph Laplacian, the eigengap may be quite small compared to the spectral radius, which can inhibit convergence rates.


This paper introduces \emph{stochastic parallelizable eigengap dilation} or \ack{}, an approach to accelerate eigendecompositions, particularly for the learning of embeddings for clustering large graphs. \ack{} improves convergence rates of bottom-$k$ eigendecompositions for large graphs by manipulating the graph Laplacian with cheap eigenvector-preserving transformations during eigenvector computation.

Our primary contribution is the development of an approach that favorably transforms the spectrum of the graph Laplacian in a way that 1) can be computed cheaply, 2) admits unbiased estimates, 3) can be parallelized, and 4) critically, maintains the original eigenvectors along with their rank.

In Section~\ref{sec:background}, we review the graph Laplacian and its connection to clustering via relaxation of min-cut from combinatorial graph theory. Next, Section~\ref{sec:background_evp} gives background on computing eigenvectors of large matrices with special attention to computing eigenvectors of graph Laplacians. We then describe our approach, \ack{}, in Section~\ref{sec:method}. Section~\ref{sec:experiments} then evaluates the viability of \ack{} by examining it in the non-stochastic, sequential setting on several different domains: clustering large graphs, clustering large graphs with uncertain edges filled in by \emph{link prediction}, and computing proto-value functions for use in reinforcement learning. Section~\ref{sec:conclusion} concludes. Appendix A presents additional experiments and Appendix~\ref{sec:related} reviews related work.


\section{Spectra of Graph Laplacians}\label{sec:background}
The Laplacian of a graph $\mathcal{G} = (\mathcal{V}, \mathcal{E})$ with node set (vertices) $\mathcal{V}$ and edge set $\mathcal{E}$ is given by $L = D-A$ where $D$ is a diagonal matrix containing the degree of each node and $A$ is the adjacency matrix of the graph $\mathcal{G}$. The graph Laplacian can also be represented as $L = X^\top X = \sum_{e \in \mathcal{E}} x_e x_e^\top$ where the incidence matrix $X$ encodes edges on its rows. Each row $x_e \in \mathbb{R}^{\vert \mathcal{V} \vert}$ corresponds to an edge $(i, j$) and contains two nonzero entries: a $1$ at index $\min(i, j)$ and a $-1$ at index $\max(i, j)$. Note that $L \mathbf{1} = \sum_{e \in \mathcal{E}} x_e x_e^\top \mathbf{1} = \sum_{e \in \mathcal{E}} x_e (1 - 1) = \mathbf{0}$ implying the ones vector is an eigenvector of $L$ with zero eigenvalue. We denote eigenvalues in increasing order by $\lambda_i$ where $\lambda_i \le \lambda_{i+1}$.

\subsection{Relaxation of Min Cut}
Given the incidence matrix representation of the graph Laplacian, consider the following quadratic form~\citep{von2007tutorial}:
\begin{align}
    \texttt{cut}(S, \bar{S}) &= v^\top (X^\top X) v = \sum_{e \in \mathcal{E}} (x_e^\top v)^2 \nonumber
    \\ &= \sum_{e=(i,j) \in \mathcal{E}} (v_i - v_j)^2 \label{eq:num_cut}
\end{align}
where $S$ is a set of nodes, $\bar{S}$ is its complement, and $v$ is a vector of length number of nodes that indicates whether a node $i$ belongs to $S$ ($v_{i \in S}=1$) or not $(v_{i \not\in S}=-1)$. By inspection, \eqref{eq:num_cut} counts (4$\times$) the number of edges crossing between $S$ and its complement $\bar{S}$. 

Finding $v$ such that the \emph{cut} is minimized and $\vert v_i \vert = 1$ for all $i$ is NP-Hard. Relaxing the feasible set to include all $v$ such that $||v||^2_2 = \vert \mathcal{V} \vert$ results in an  eigenvalue problem, in which case, $||v||^2_2 = \vert \mathcal{V} \vert$ can be replaced with $||v||^2_2 = 1$ w.l.o.g.:
\begin{align}
    \min_{v^\top 1 = 0} v^\top L v \quad \text{s.t.} \quad v^\top v = 1 \label{eq:num_cut_relax}
\end{align}
where $v^\top \mathbf{1} = 0$ is included to avoid the trivial solution of assigning all nodes to the same cluster. Note the solution to~\eqref{eq:num_cut_relax} is the 2nd smallest eigenvector of $L$, also called the Fiedler vector. To assign hard clusters, we can then either threshold $v$ or run $k$-means with $k=2$ on $v$.

Cheeger's inequality bounds a related objective called the \emph{normalized cut} in terms of the eigenvalues of the graph Laplacian. Let
\begin{align}
    \phi_{\mathcal{G}}(S) &= \frac{\text{cut}(S, \bar{S})}{\text{vol}(S)}
\end{align}
where vol$(S)$ denotes the number of edges incident to nodes in $S$. And define the best cut of the graph to be the $S$ (and $\bar{S}$) that solves
\begin{align}
    \rho_{\mathcal{G}} &= \underset{S \subset \mathcal{V}}{\min} \max\{ \phi_{\mathcal{G}}(S), \phi_{\mathcal{G}}(\bar{S}) \}.
\end{align}
 Then
\begin{align}
    \frac{\lambda_2}{2} \leq \rho_{\mathcal{G}} \leq \sqrt{2 \lambda_2}
\end{align}
where $\lambda_2$ is the 2nd smallest eigenvalue of $L$.

Therefore, as the number of edges in the graph increases (denominator of $\phi_{\mathcal{G}}(S)$) relative to the number of edges crossing the cut (numerator), i.e., in a ``well-clustered'' graph, $\lambda_2$ shrinks.

The results and connections discussed so far have been generalized to the setting where the graph is partitioned into $k>2$ sets~\citep{lee2014multiway}. In this case, the bottom-$k$ eigenvectors of the graph Laplacian are particularly useful for clustering. Similarly to before,
\begin{align}
    \frac{\lambda_k}{2} \leq \rho_{\mathcal{G}}(k) \leq c k^2 \sqrt{\lambda_k}
\end{align}
for some $c > 0$ with
\begin{align}
    \rho_{\mathcal{G}}(k) &= \underset{S_1,S_2,\ldots,S_k \subset \mathcal{V}}{\min} \max\{ \phi_{\mathcal{G}}(S_i): i = 1, 2, \ldots k \}
\end{align}
where $\{S_i\}$ represent all possible $k$-way partitions of the graph. Similar reasoning still applies \textemdash if there are $k$ clear clusters in the graph, we should expect $k$ eigenvalues $\ll 1$.

In addition, if the graph edges are weighted, the above analysis is extended by summing edge weights rather than counting edges ($w_{ij} = 1$). The graph Laplacian can be written more generally as $X^\top W X$ where $W$ is a diagonal matrix containing the edge weights.

\section{Eigendecomposition}\label{sec:background_evp}


The incidence matrix representation of the graph Laplacian makes it clear the matrix is positive semi-definite. It also makes clear its connection to the singular value decomposition given the fact that the eigenvectors of $X^\top X$ are the same as the right singular vectors of $X$. This means various algorithms for SVD can be applied to spectral clustering.


Most scalable SVD methods focus on finding the top-$k$ eigenvectors of $X^\top X$. To find the bottom-$k$ eigenvectors, a common approach uses 
\begin{align}
    L^- &= \lambda^\ast I - L \label{eqn:reveig}
\end{align}
with $\lambda^\ast > \lambda_{\vert \mathcal{V} \vert}$, which turns the bottom-$k$ eigenvectors of $L$ into the top-$k$ eigenvectors of $L^-$, and allows the use of a top-$k$ solver for finding eigenvectors for spectral clustering.


The convergence of most scalable SVD methods also depends inversely on the eigengaps of the matrix. The eigengaps are the differences between consecutive eigenvalues
\begin{align}
    g_i &= \lambda_{i+1} - \lambda_{i}
\end{align}
For example, in~\cite{gemp2020eigengame}, the number of steps to convergence depends on the ratio $\frac{\lambda_{\vert \mathcal{V} \vert}}{g_i}$ for each consecutive eigenvector $i$ (larger ratio means more steps). Intuitively, algorithms struggle to rank eigenvectors when they have similar eigenvalues (small eigengaps), especially in the stochastic setting where eigenvalues / eigenvectors are estimated from random samples of the data. In the large data setting where the entire dataset cannot be stored in memory, approaches that sample data (in this case batches of edge vectors $x_e$) are often all that are viable.

\section{Method}\label{sec:method}

In this section, we describe our approach to accelerating the convergence of solvers in finding the bottom-$k$ eigenvectors of well-clustered graphs. We are particularly interested in constructing an approach that can scale to large graphs. As suggested in the previous section, we aim to improve convergence rates by dilating (enlarging) the eigengaps of the graph Laplacian relative to its spectral radius. If successful, this should reduce the number of steps required for convergence. 



\subsection{Eigenvector Preserving Transformations}
Any monotonically increasing polynomial function of a matrix preserves eigenvectors and their rank. If we know we only want to discover eigenvectors with eigenvalues below some threshold $\lambda_c$, the function $f$ actually only needs to be monotonic below this threshold; it can then be non-monotonic above $\lambda_c$ as long as $f(x) > f(\lambda_c)$ for $x > \lambda_c$. For example, in the spectral clustering setting when using the normalized graph Laplacian, we might only be interested in eigenvectors with eigenvalues below $\lambda_c=1$, i.e., clusters with fewer edges crossing between clusters than edges within a cluster.

Once we have a monotonically increasing function $f$ that is tailored to dilate the eigengaps of the problem at hand, we will reverse the spectrum with~\eqref{eqn:reveig} to compute the bottom-$k$ eigenvectors. This is the approach we take in experiments in Section~\ref{sec:experiments}.

In order to dilate the eigengaps of small eigenvalues relative to those of large eigenvalues, we require nonlinear operators. Naively, one could leverage the eigendecomposition of $L$ and compute, for example $e^L$ as
\begin{align}
    e^L &= e^{V \Lambda V^\top} = V e^{\Lambda} V^\top = V \texttt{diag}(e^{\lambda_i}) V^\top.
\end{align}

But obviously, this requires the eigendecomposition which we do not have and are trying to compute.

\subsection{Series Expansions of Matrix Transformations}\label{sec:fun_approx}
Instead, we can explore power series approximations to nonlinear matrix operators. For example, the matrix exponential is given by 
\begin{align}
    e^L &= \sum^\infty_{i=0} \frac{L^i}{i!}.
\end{align}
This operation dilates eigengaps for large eigenvalues, but using a similar series we can compute $-e^{-L}$ which shrinks large eigenvalues relative to small ones. By shrinking large eigenvalues relative to smaller ones, the scale of the spectrum is reduced. Therefore, the ratio $\frac{\lambda_{|\mathcal{V}|}}{g_i}$ is smaller, and convergence is sped up. Note that the maximum eigenvalue under this transformation is $<0$ (i.e., we can set $\lambda^* = 0$ in~\eqref{eqn:reveig}) and the spectral radius is $1$.


\subsection{Scaling: Stochastic \& Parallel}\label{sec:scaling}
\label{powers_of_incidence_as_rand_walk}

Recall that the incidence matrix $X$ encodes edges on its rows and that the graph Laplacian can be written as $L = X^\top X = \sum_e x_e x_e^\top$. Therefore, any power $\ell$ of the graph Laplacian can be written as
\begin{align}
    L^\ell &= (\sum_{e_1 \in E} x_{e_1} x_{e_1}^\top) \ldots (\sum_{e_\ell \in E} x_{e_\ell} x_{e_\ell}^\top) \nonumber
    \\ &= \sum_{e_1 \in E} \ldots \sum_{e_\ell \in E} (x_{e_1} x_{e_1}^\top) \ldots (x_{e_\ell} x_{e_\ell}^\top) \nonumber
    \\ &= \sum_{e_1 \in E} \ldots \sum_{e_\ell \in E} \underbrace{\Big[ \prod_{j=1}^{\ell-1} x_{e_j}^\top x_{e_{j+1}} \Big]}_{\alpha_c} x_{e_1} x_{e_\ell}^\top \nonumber
    \\ &= \sum_{c_1 \in E^\ell} \alpha_c x_{e_1} x_{e_\ell}^\top \label{eq:walk_summand}
\end{align}

where $E^\ell$ represents the Cartesian product of the edge set. Note that the scalar, denoted by $\alpha_c$, is only non-zero if each of its constituent inner products $x_{e_j}^\top x_{e_{j+1}}$ is non-zero. And the inner product of two edge vectors is only non-zero if those two edges are incident. Table~\ref{tab:alpha_c} lists the possible values for inner products of pairs of edges assuming $x_e$ is encoded such that $x_{e_i} = 1$ and $x_{e_j} = -1$ if $i < j$ where $e = (i, j)$.
\begin{table}[]
    \centering
    \begin{tabular}{c|c|c}
        disconnected edges & $i \rightarrow j , k \rightarrow l$ & 0
        \\ serial & $i \rightarrow j \rightarrow l$ & -1
        \\ converging & $i \rightarrow j \leftarrow l$ & 1
        \\ diverging & $i \leftarrow j \rightarrow l$ & 1
        \\ repeated & $i \Rightarrow j$ & 2
    \end{tabular}
    \caption{Recall that edge directions are defined by node numbers, not by the direction of a walk on the graph.}
    \label{tab:alpha_c}
\end{table}

In other words, $\alpha_c$ is only non-zero if the sequence of edges $e_1 \ldots e_\ell$ forms a chain (hence the subscript $c$ of $\alpha_c$). Therefore, computing a power of the graph Laplacian is equivalent to computing a weighted sum of outer products over length-$\ell$ walks in the \emph{edge incidence graph}\footnote{By \emph{edge incidence graph}, we mean to define a new undirected graph whose nodes represent the edges in the original graph and whose edges indicate whether two edges (now nodes) are incident. Note every node in the new graph also has a self-loop edge.}.

This suggests a highly parallelizable method. Assuming access to $d$ graph ``walkers``, let each walker, in parallel, sample a node and conduct a random walk of length-$\ell$ on the graph. Average each walker's calculation of the summand in~\eqref{eq:walk_summand} to estimate $\frac{1}{\vert \mathcal{E} \vert^\ell} L^\ell$. Note that if each walk is unbiased, then the average is unbiased.


In order to sample a walk uniformly at random, one could use rejection sampling. Pick an edge in the graph uniformly at random. Then perform a random walk on the edge incidence graph keeping track of the degree of every node visited in the edge incidence graph. Let $\texttt{deg}(e_i)$ be the degree of node $i$ in the edge incidence graph. Then the probability of a given length-$\ell$ walk is
\begin{align}
    p_\ell &= \frac{1}{\vert \mathcal{E} \vert} \prod_{i=1}^\ell \frac{1}{\texttt{deg}(e_i)}.
\end{align}
Also, let $\texttt{deg}^*$ be an upper bound on the maximum degree of any node in the original graph. Then $\texttt{deg}^*_{inc} = 2 \texttt{deg}^* - 1$ is an upper bound on the degree of any node in the edge incidence graph. Let
\begin{align}
    p_{min} = \frac{(\texttt{deg}^*_{inc})^{-l}}{\vert \mathcal{E} \vert}. 
\end{align}
Finally, if walks are rejected with probability $\frac{p_{min}}{p_\ell}$, then all walks in the edge incidence graph will occur with probability $p_{min}$ ensuring uniform sampling.

Linearity of expectation says you can use same length-$\ell$ walk for unbiased estimates of all shorter walks because $\mathbb{E}[\sum_i \gamma_i L^i] = \sum_i \gamma_i \mathbb{E}[L^i]$. This means $L^i$ and $L^j$ can be correlated. Therefore, to generate a single unbiased estimate for all $L^i$ with $1 \le i \le \ell$, simply continue performing random walks of length $\ell$ until at least one \emph{sub}walk of each length less than or equal to $\ell$ has passed the rejection step. Note that with sufficient compute a large batch of random walks can be generated in parallel to reduce runtime. Furthermore, the generation of a single random walk can be parallelized as well~\citep{lkacki2020walking,kapralov2021efficient}.

In general, a product of $\ell$ matrices can be computed in parallel via a binary tree in $\log(\ell)$ time but here we take advantage of the structure of $L$ to compute it in a different way. We mention this as our approach can generalize beyond graph Laplacians given general knowledge of the spectrum




\section{Experiments}\label{sec:experiments}
We compute the bottom-$k$ eigenvectors of matrices with spectra that have several eigenvalues less than $1$. This is meant to simulate conditions common in well separated graphs and clustering problems~\citep{peng2015partitioning}. The transformations we consider here are not robust to all graph types. Different classes of graphs will benefit from different transformations, but with prior knowledge of the problem setting and spectrum of the graph Laplacian, it should be possible to generalize our approach to other settings.

\subsection{Manipulating Spectra of Graph Laplacians}
In this paper, we propose manipulating the spectrum of graph Laplacians in order to reduce the ratio $\frac{\lambda_{\vert \mathcal{V} \vert}}{g_i}$ especially for the bottom-$k$ eigenvalues. The goal is to transform the Laplacian in such a manner that larger eigengaps are shifted to the bottom of the eigenvalue spectrum. If we then only compute the bottom-$k$ eigenvectors we can potentially reduce computation significantly, especially if we use cheap approximations of the transformations as discussed in Section~\ref{sec:fun_approx}.

In the following experiments, we compute both exact and series approximations to nonlinear transformations of the graph Laplacian and observe its effect on convergence of two representative, iterative SVD solvers: Oja's algorithm~\citep{shamir2015stochastic} and $\mu$-EG~\citep{gemp2021eigengame}.

\begin{table}[]
    \centering
    \begin{tabular}{c|c}
        Matrix Logarithm (add $\epsilon \ll 1$) & $\log(L + \epsilon)$ \\
        Taylor Series  of $\log(L + \epsilon I)$ & $\sum_{i=1}^\ell (-1)^{i+1} \frac{(L + \epsilon I - I)^i}{i}$ \\
        Negative Decaying Exponential & $-e^{-L}$ \\
        Taylor Series  of $-e^{-L}$ & $-\sum_{i=0}^\ell \frac{(-L)^i}{i!}$ \\
        Limit Approximation of $-e^{-L}$ & $-(I - L/\ell)^\ell$ \\
    \end{tabular}
    \caption{Transformation Functions ($\ell$ is odd)}
    \label{tab:transf_funs}
\end{table}

\subsection{Evaluating Convergence}
We consider two types of evaluations: normalized subspace error and longest eigenvector streak.

\paragraph{Subspace Error} We use the definition of subspace error in~\citep{tang2019exponentially, gemp2020eigengame}:
\begin{align}
    \delta^t &= 1 - \frac{1}{k} \text{tr}(U^\ast P_t)
\end{align}
where $U^\ast = V^* V^{*\top}$ is the ground truth orthogonal projector of the principal subspace and $P^t = V V^\dagger$ the orthogonal projector of the approximation of the subspace at time $t$.

\paragraph{Eigenvector Streak} We use the definition in~\citep{gemp2020eigengame}, where we measure the number of consecutive principal components that are within some $\epsilon$ of the ground truth principal components. This metric is harsher than subspace error and measures whether the actual bottom-$k$ eigenvectors are being approximated. Below we will show results on a variety of problems. For all figures except Fig.~1, the horizontal axis is plotted on a logarithmic scale to make it more convenient to interpret convergence rate differences in terms of orders of magnitude.


\subsection{Proto-value Functions for MDPs}\label{sec:pvf}
\begin{figure}
    \centering
    \includegraphics[width=0.45\textwidth]{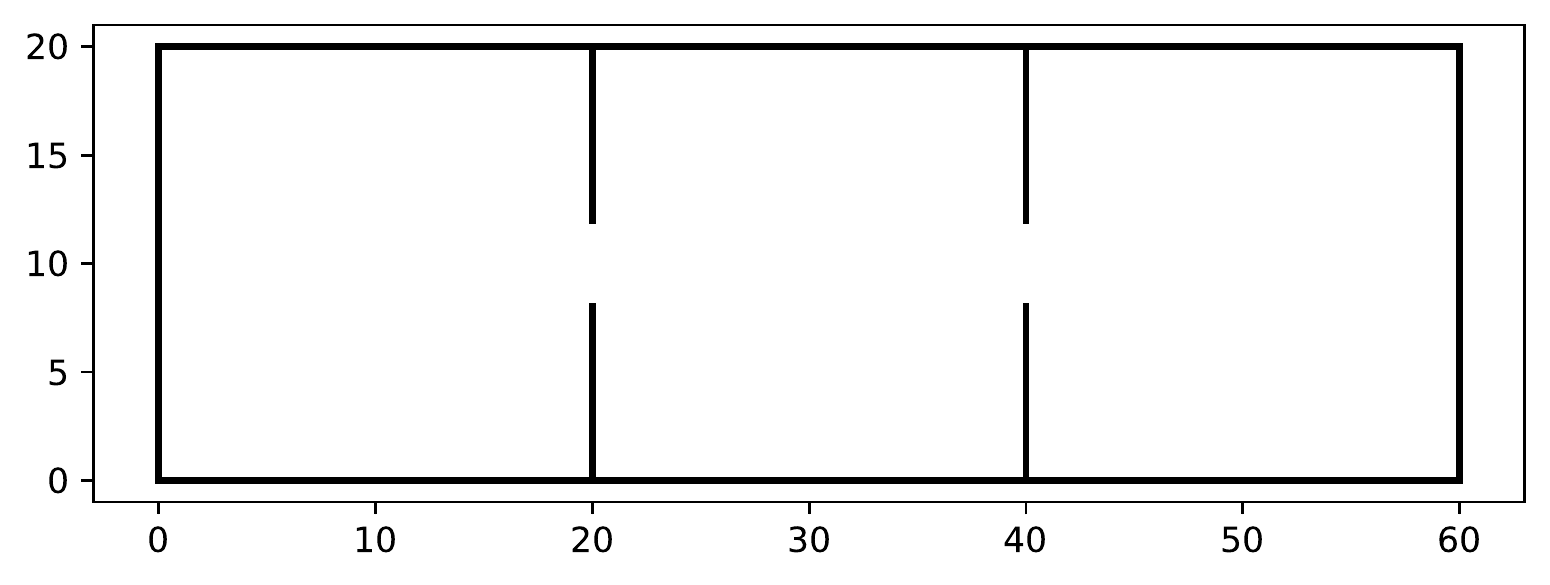}
    \caption{3-Room MDP Grid World ($s=2, h=10$).}
    \label{fig:mdp_gridworld}
\end{figure}

We consider computing the proto-value functions, basis functions for value functions of a tabular Markov Decision Process (MDP) illustrated in Figure~\ref{fig:mdp_gridworld}. Specifically, we consider an MDP with 3 consecutive rooms with the middle connected to each of the outer rooms by small doors. The grid world is $10s + 1$ cells tall and $30s + 1$ cells wide. The doorways take up $\frac{1}{h}$ of the available vertical space ($\frac{10s+1}{h}$ cells tall).

The top-$k$ proto-value functions are computed by first constructing the graph of all possible state transitions in the MDP (states $s$ are nodes and undirected edges indicate possible transitions $s$$\rightarrow$$s'$; note $s$$\leftarrow$$s'$ is also included in the graph) and then computing the bottom-$k$ eigenvectors of the corresponding graph Laplacian. Proto-value functions have been shown to be useful as a basis for learning value functions in reinforcement learning tasks~\citep{mahadevan2005proto}.

\begin{figure}
    \centering
    \includegraphics[width=0.225\textwidth]{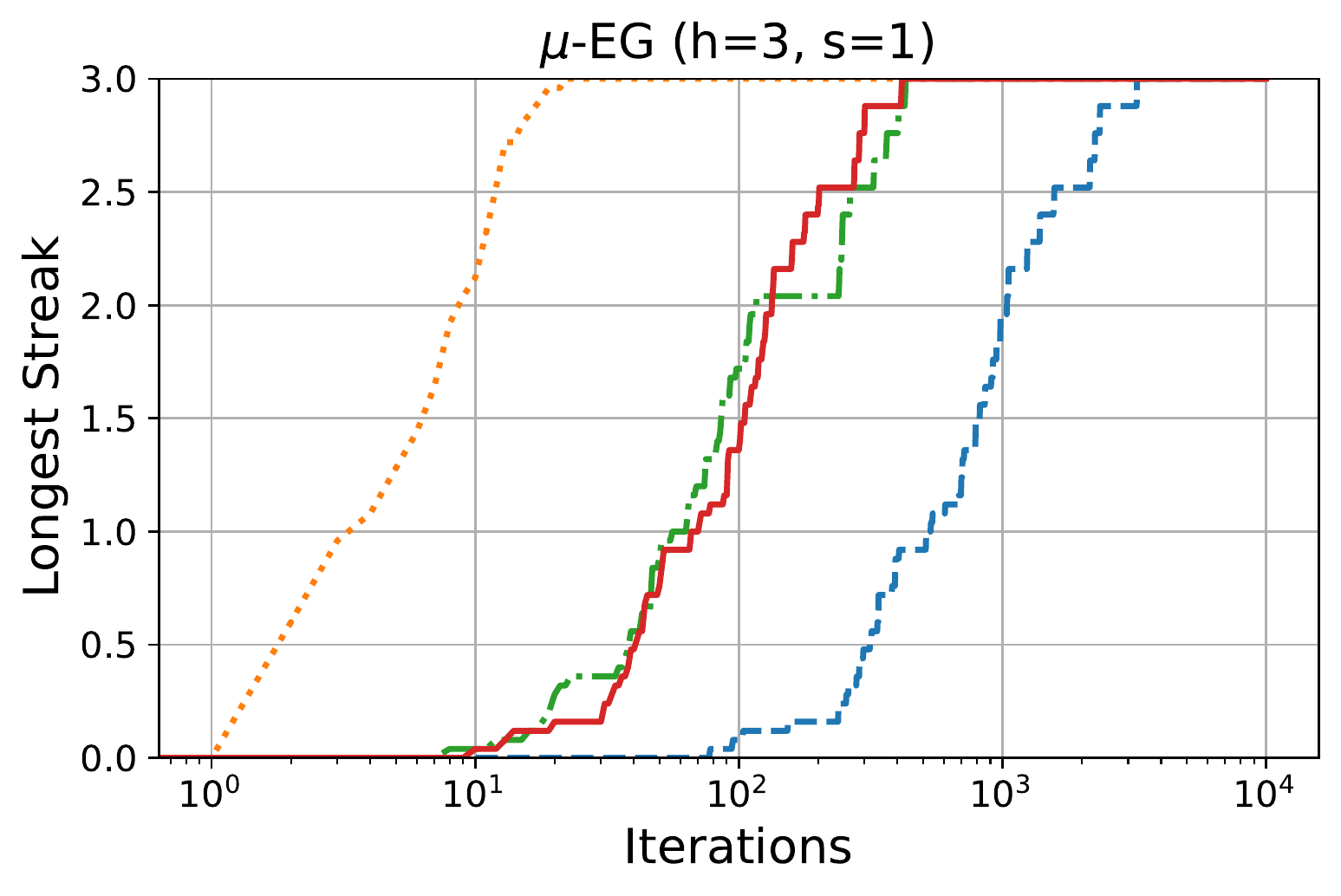}
    \includegraphics[width=0.225\textwidth]{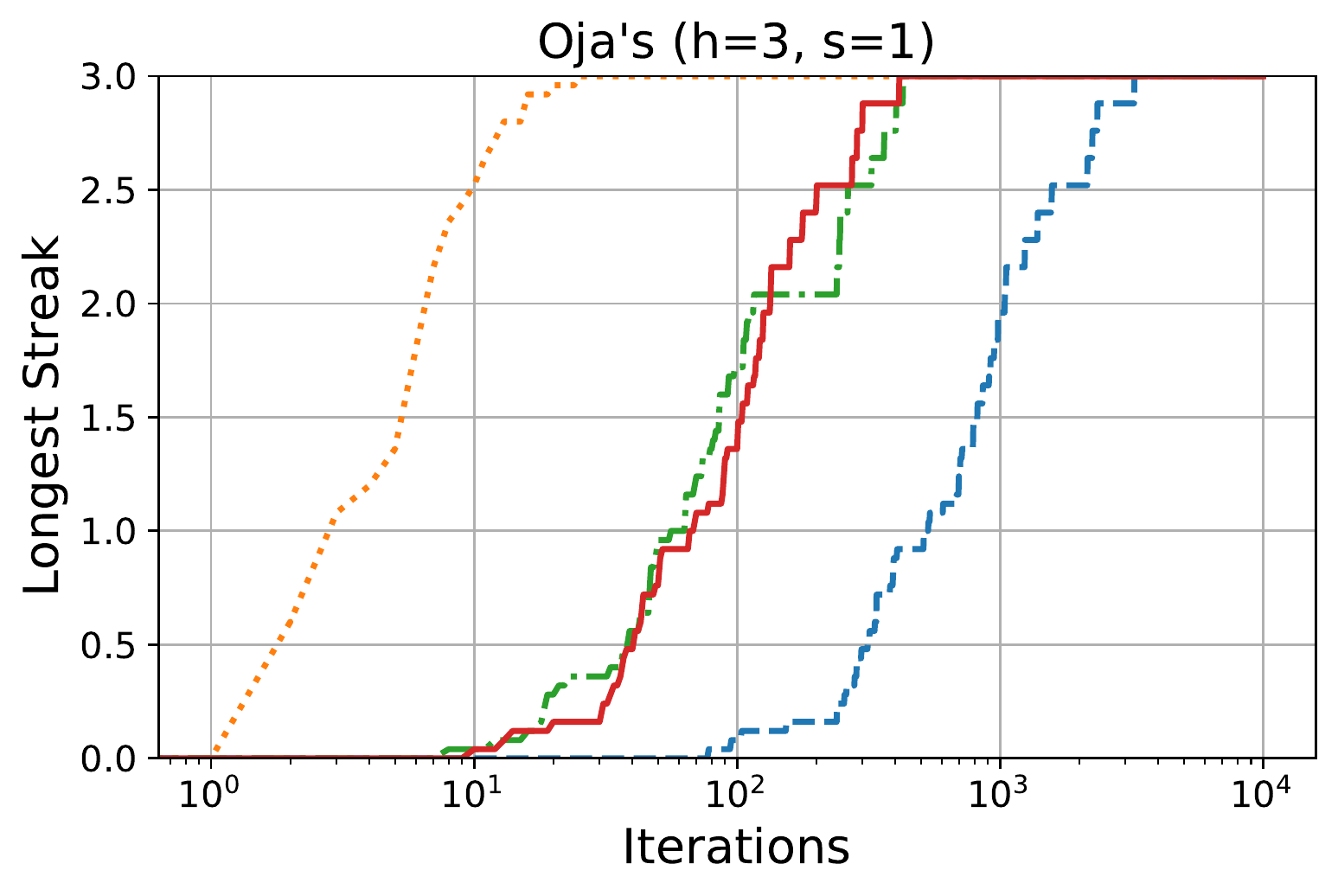}
    \includegraphics[width=0.225\textwidth]{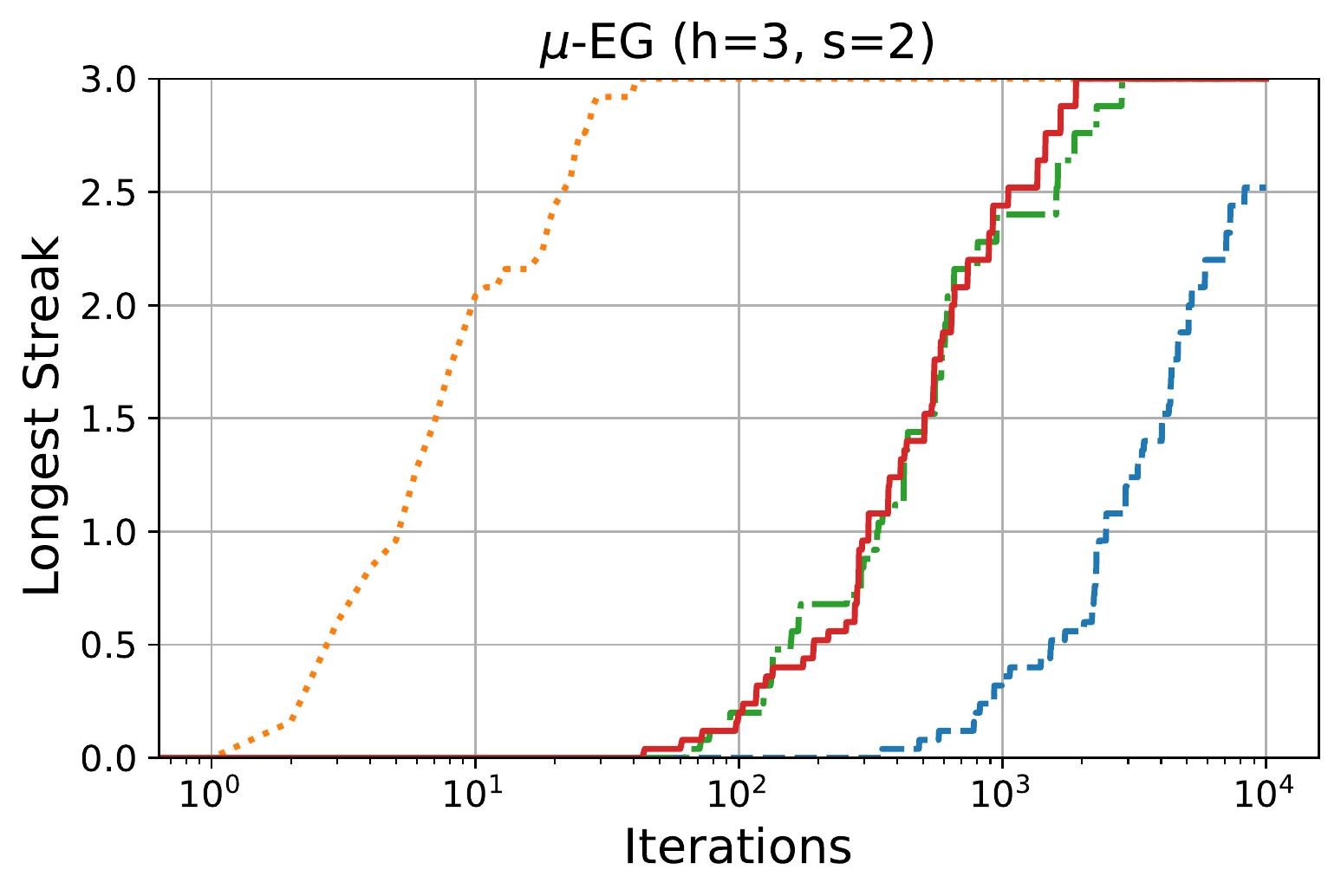}
    \includegraphics[width=0.225\textwidth]{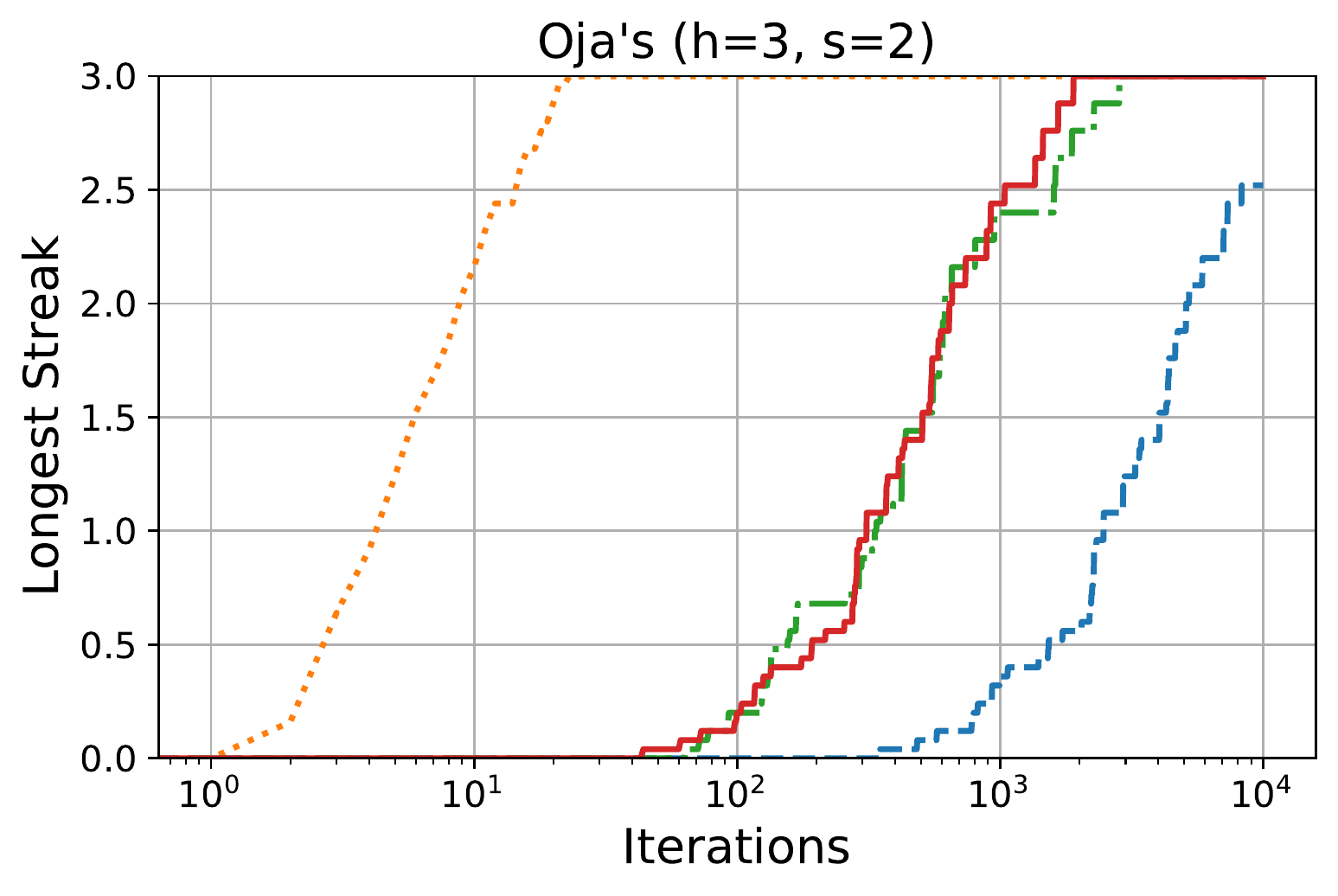}
    \includegraphics[width=0.225\textwidth]{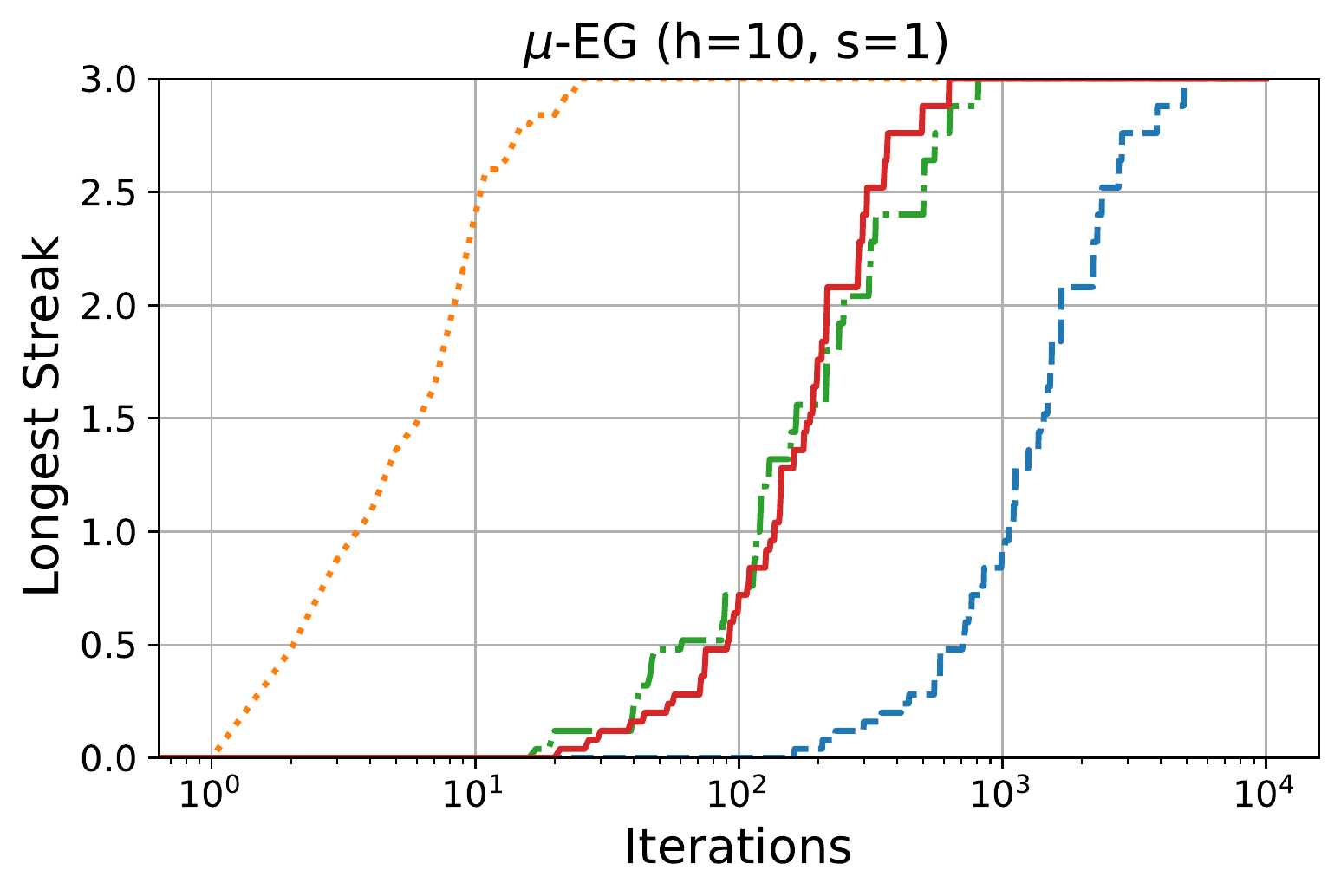}
    \includegraphics[width=0.225\textwidth]{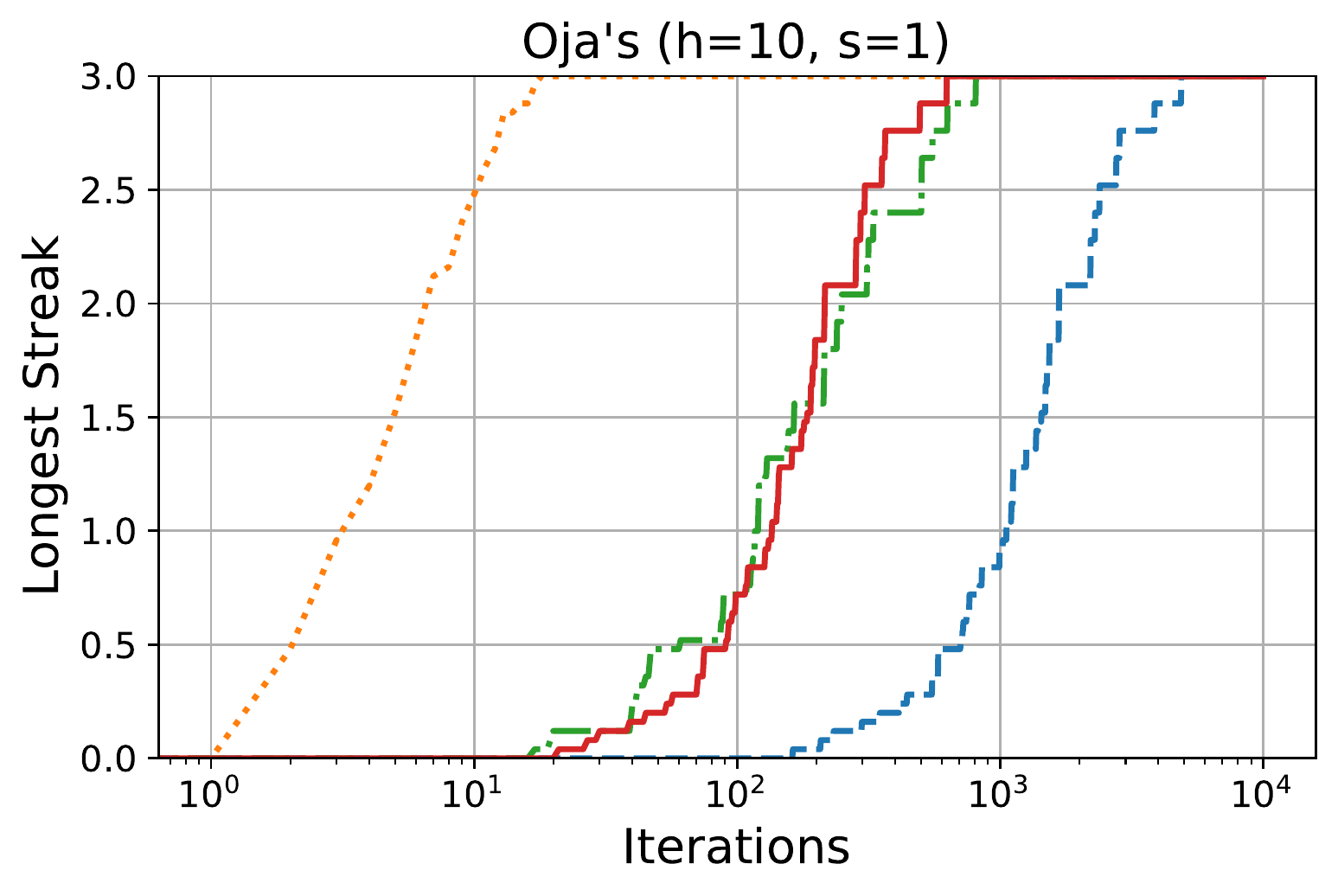}
    \includegraphics[width=0.225\textwidth]{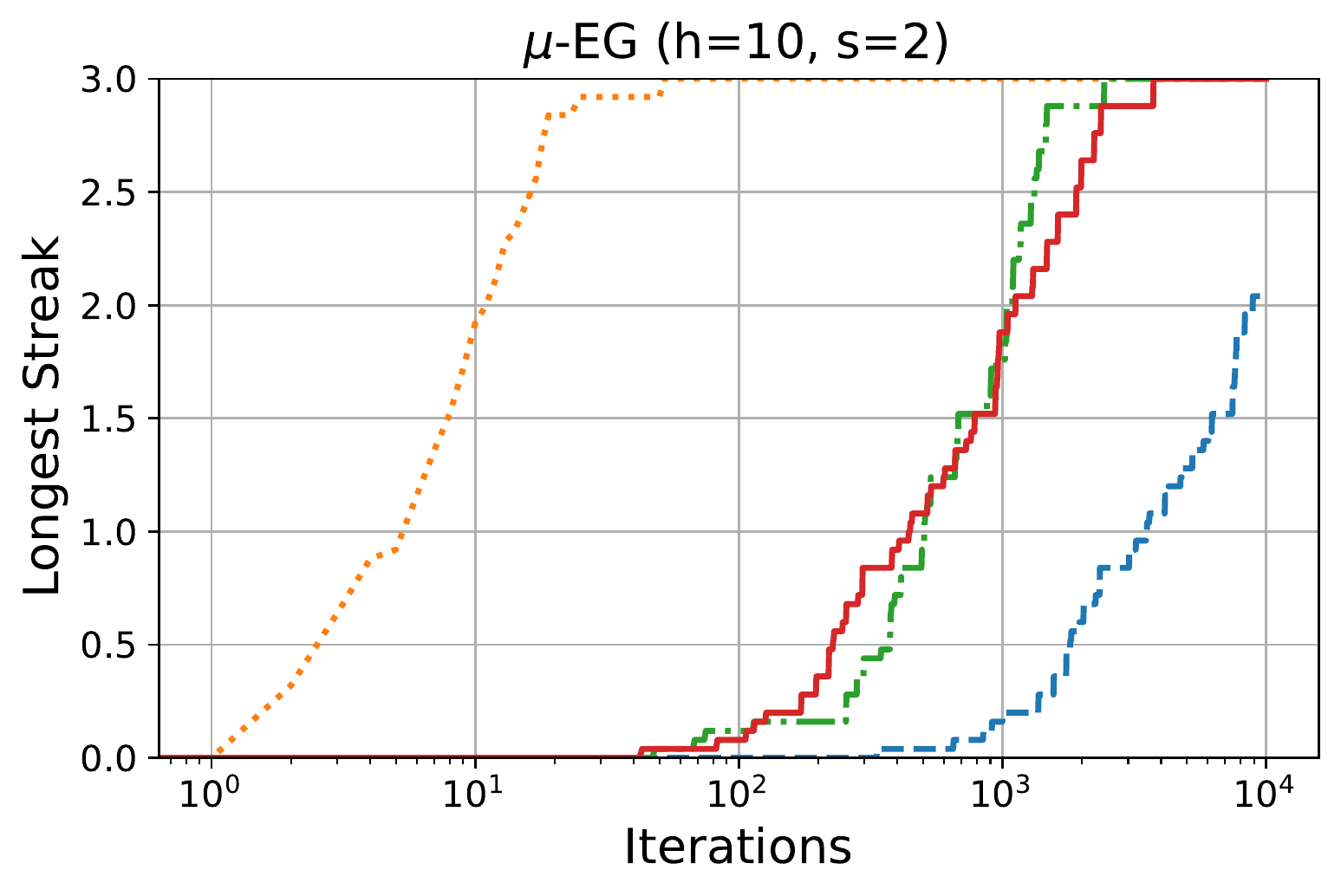}
    \includegraphics[width=0.225\textwidth]{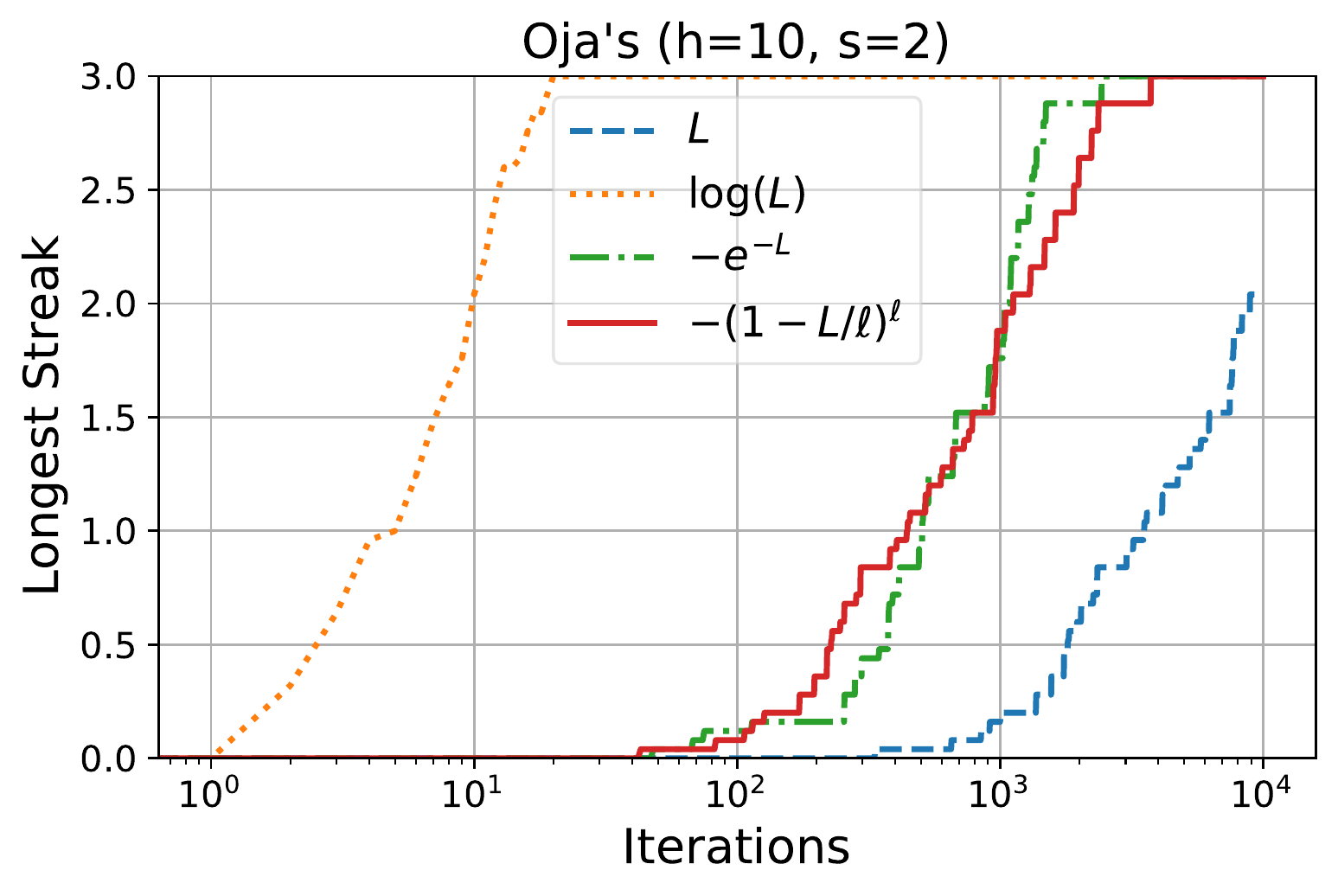}
    \caption{3-Room MDP. Longest eigenvector streak (higher is better) is plotted over training for two different scalable SVD methods: $\mu$-EG and Oja's algorithm. Three nonlinear matrix transformations of matrix $L$ are compared against the identity transformation. One transformation (in red) is a series approximation of degree $\ell=251$ to the exact operation (in green).}
    \label{fig:mdp}
\end{figure}

Figure~\ref{fig:mdp} demonstrates that certain transformations of the graph Laplacian do, in fact, accelerate convergence to the bottom-$k$ eigenvectors. Even when using a power series approximation, the number of steps can be reduced by roughly an order of magnitude (10$\times$). The exact matrix logarithm reduces the number of steps by over two orders of magnitude (100$\times$). However, it is known that the Taylor series approximation of the logarithm is only convergent for $\rho(L) < 2$, and we are unable to find a series approximation that is accurate enough over the graph Laplacian's full spectrum.


\begin{figure}
    \centering
    \includegraphics[width=0.225\textwidth]{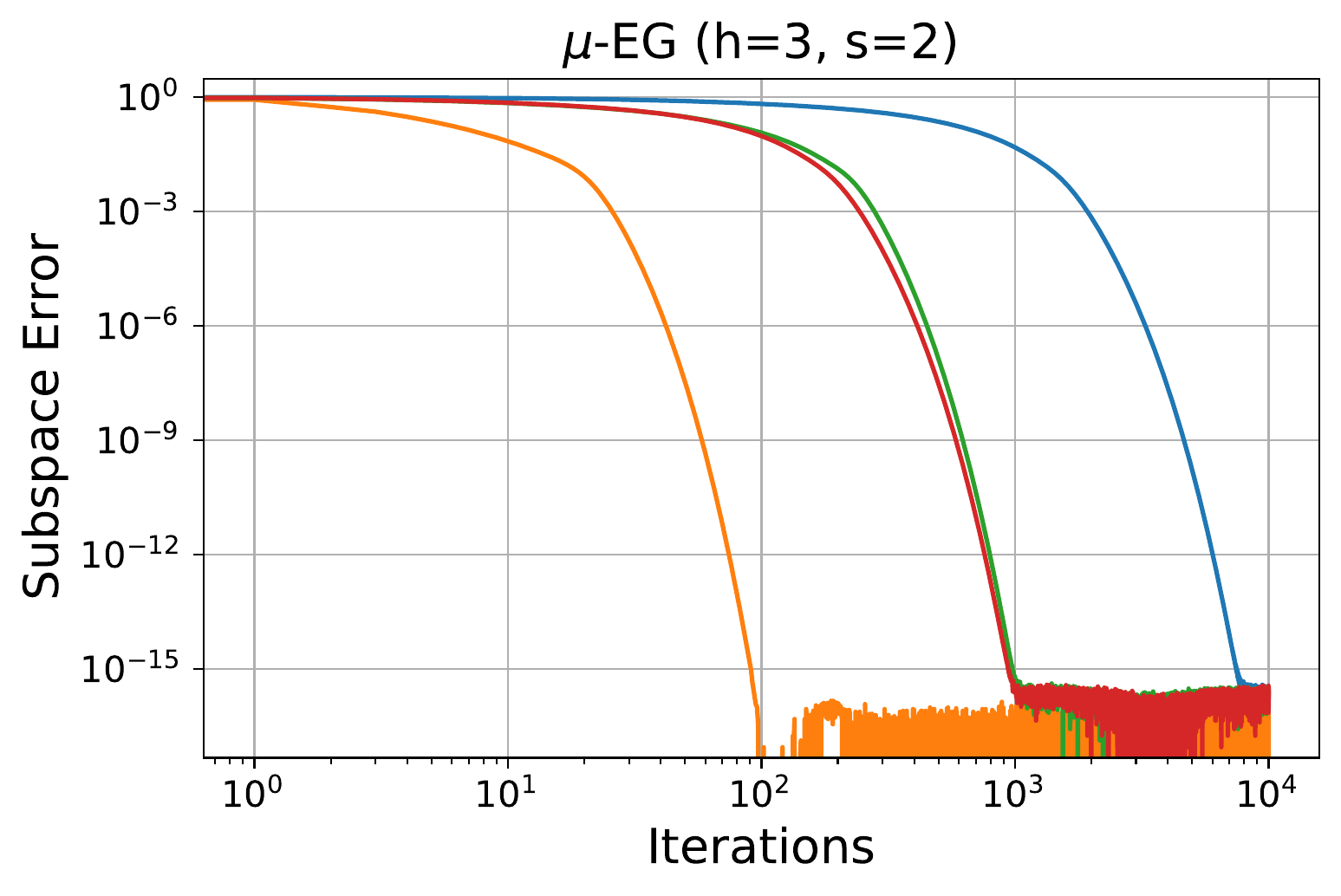}
    \includegraphics[width=0.225\textwidth]{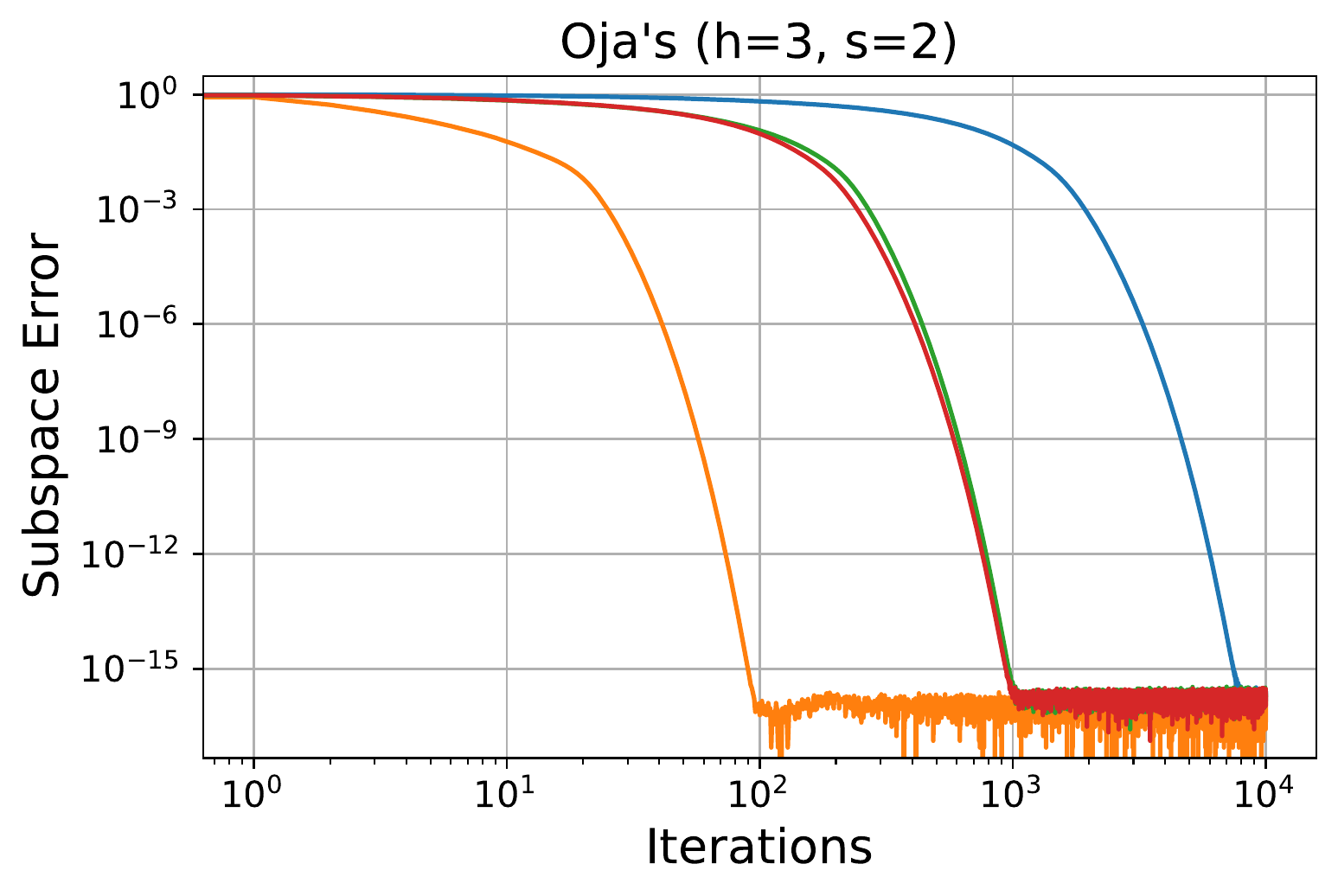}
    \includegraphics[width=0.225\textwidth]{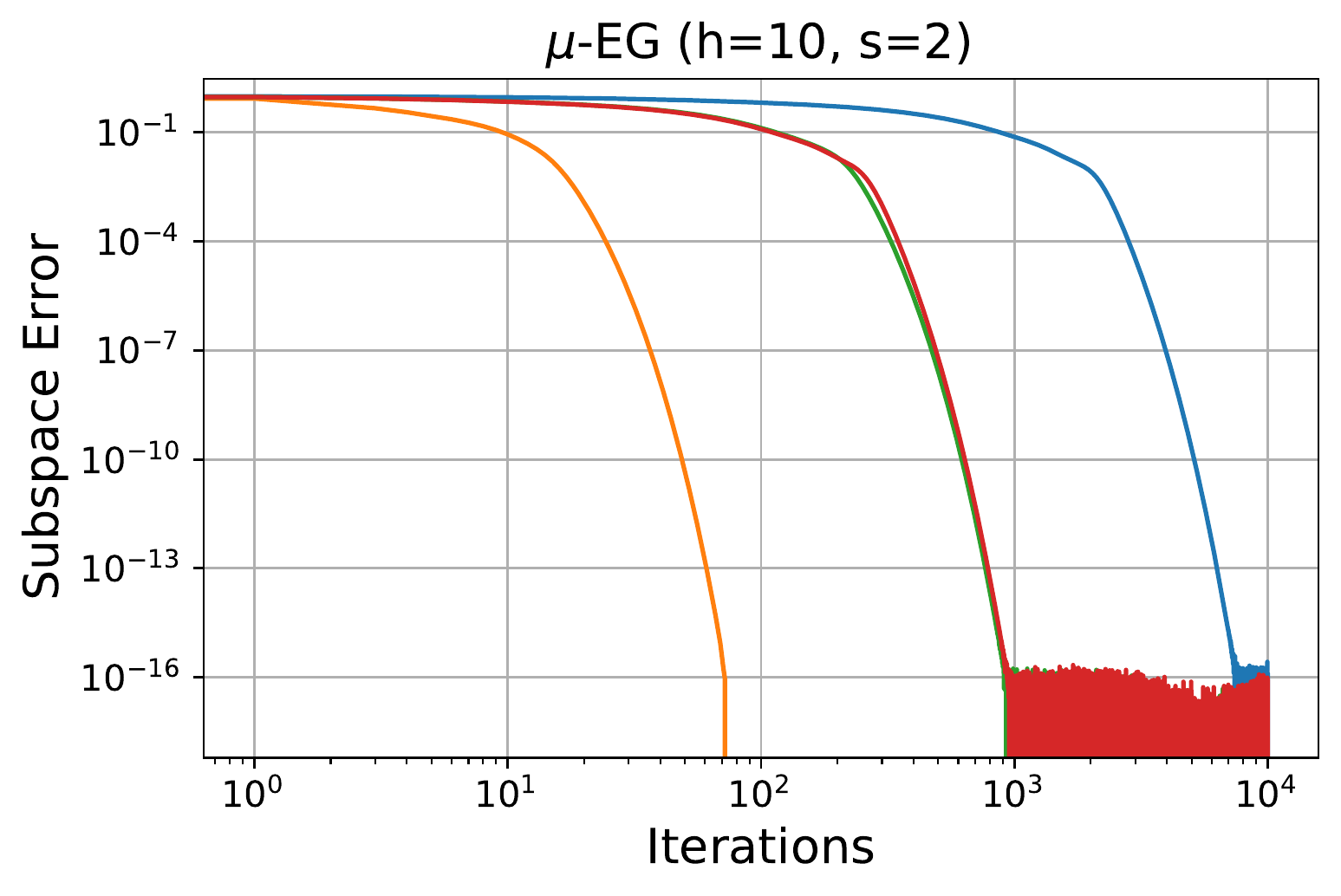}
    \includegraphics[width=0.225\textwidth]{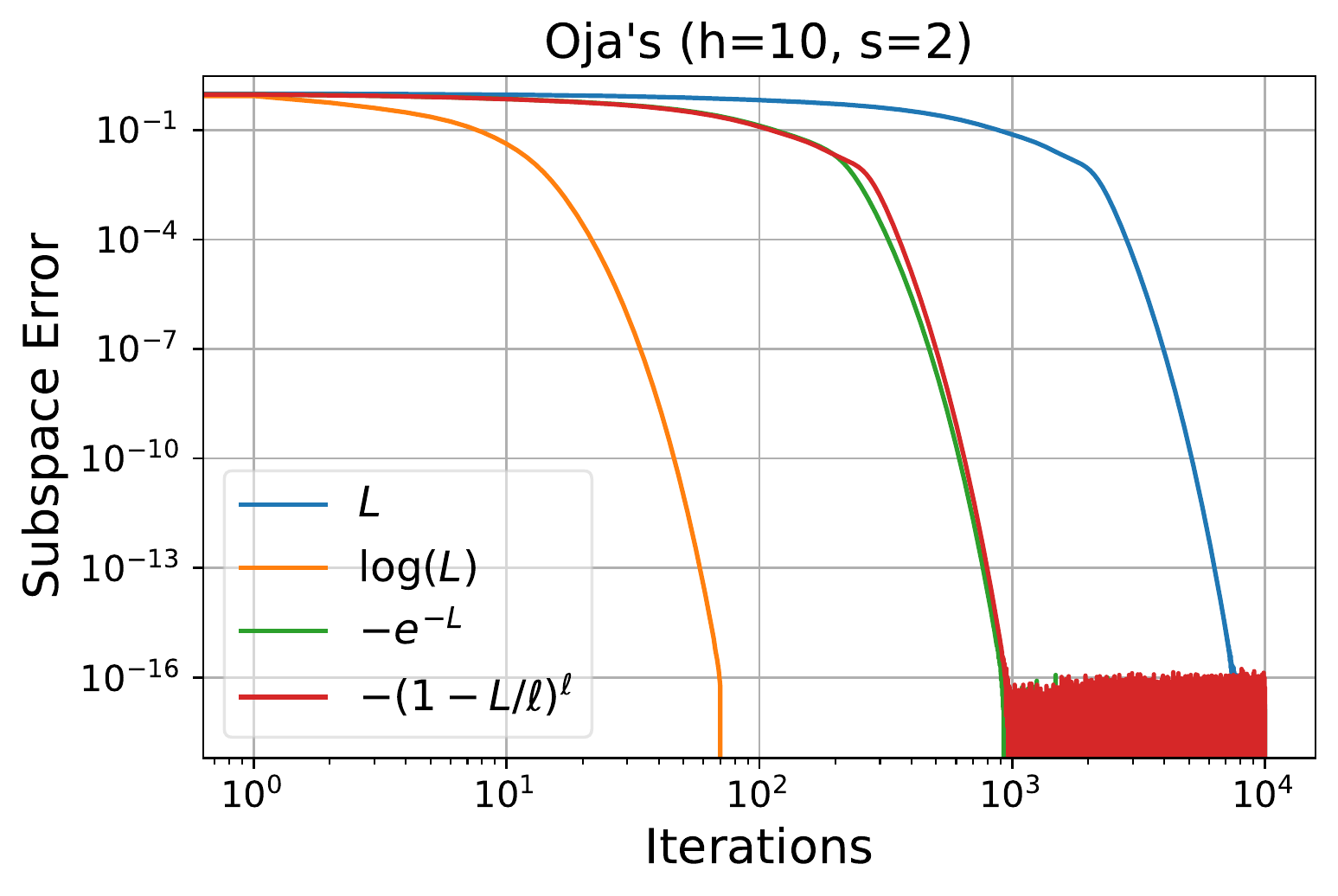}
    \caption{3-Room MDP. Subspace error (lower is better) is plotted over training for two different scalable SVD methods: $\mu$-EG and Oja's algorithm. Three nonlinear matrix transformations of matrix $L$ are compared against the identity transformation. One transformation (in red) is a series approximation of degree $\ell=251$ to the exact operation (in green).}
    \label{fig:mdp_sub}
\end{figure}

Figure~\ref{fig:mdp_sub} reveals that both the nonlinear transformations and the series approximation accelerate the minimization of subspace error as well.

\subsection{Well-Clustered Graphs}
\label{sec:cliques}

We generate graphs of $n$ nodes split into $k$ cliques. The $k$ cliques are connected to each other by a random number between $0$ and $25$ of ``short circuit'' edges between the clusters.



\begin{figure}
    \centering
    \includegraphics[width=0.225\textwidth]{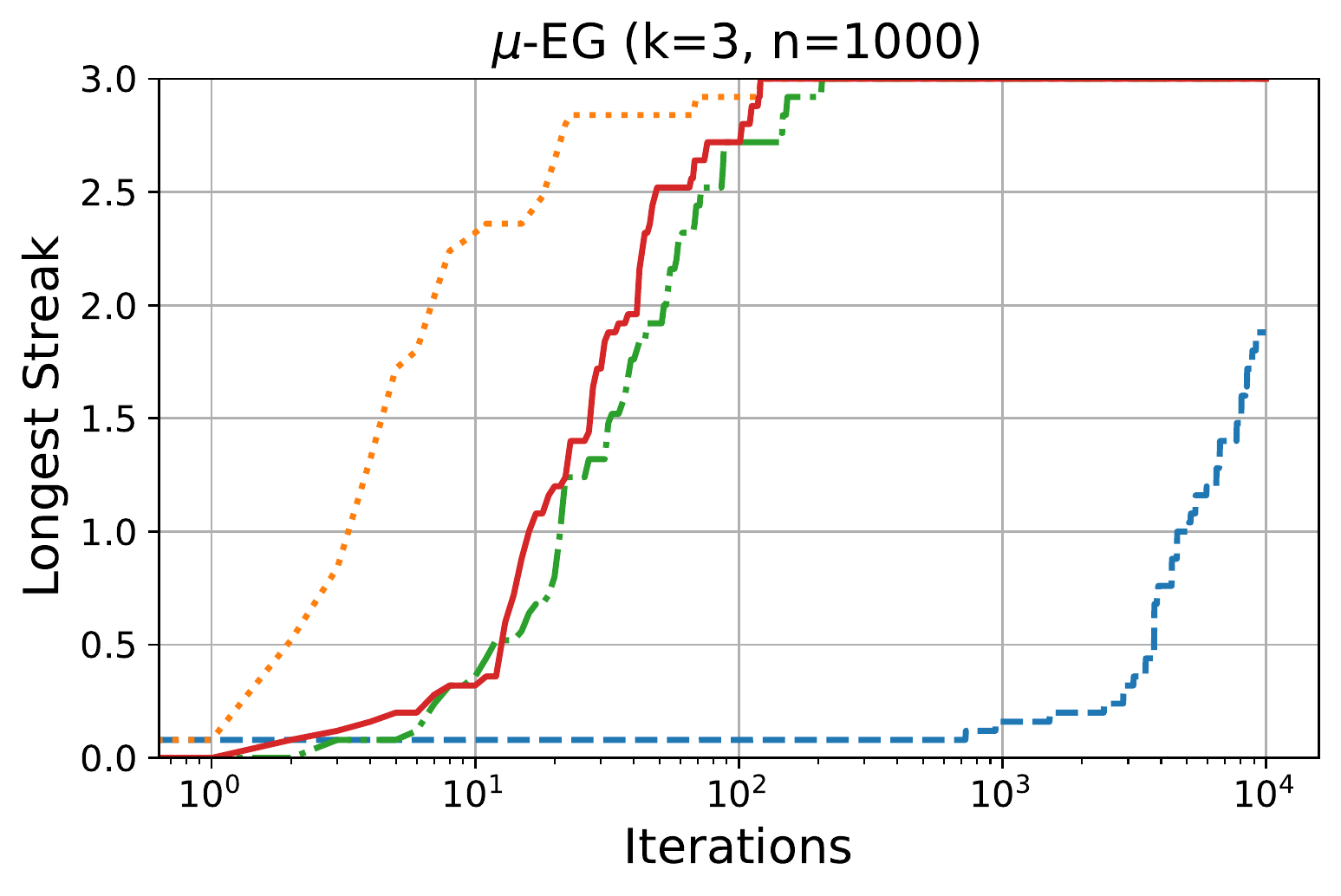}
    \includegraphics[width=0.225\textwidth]{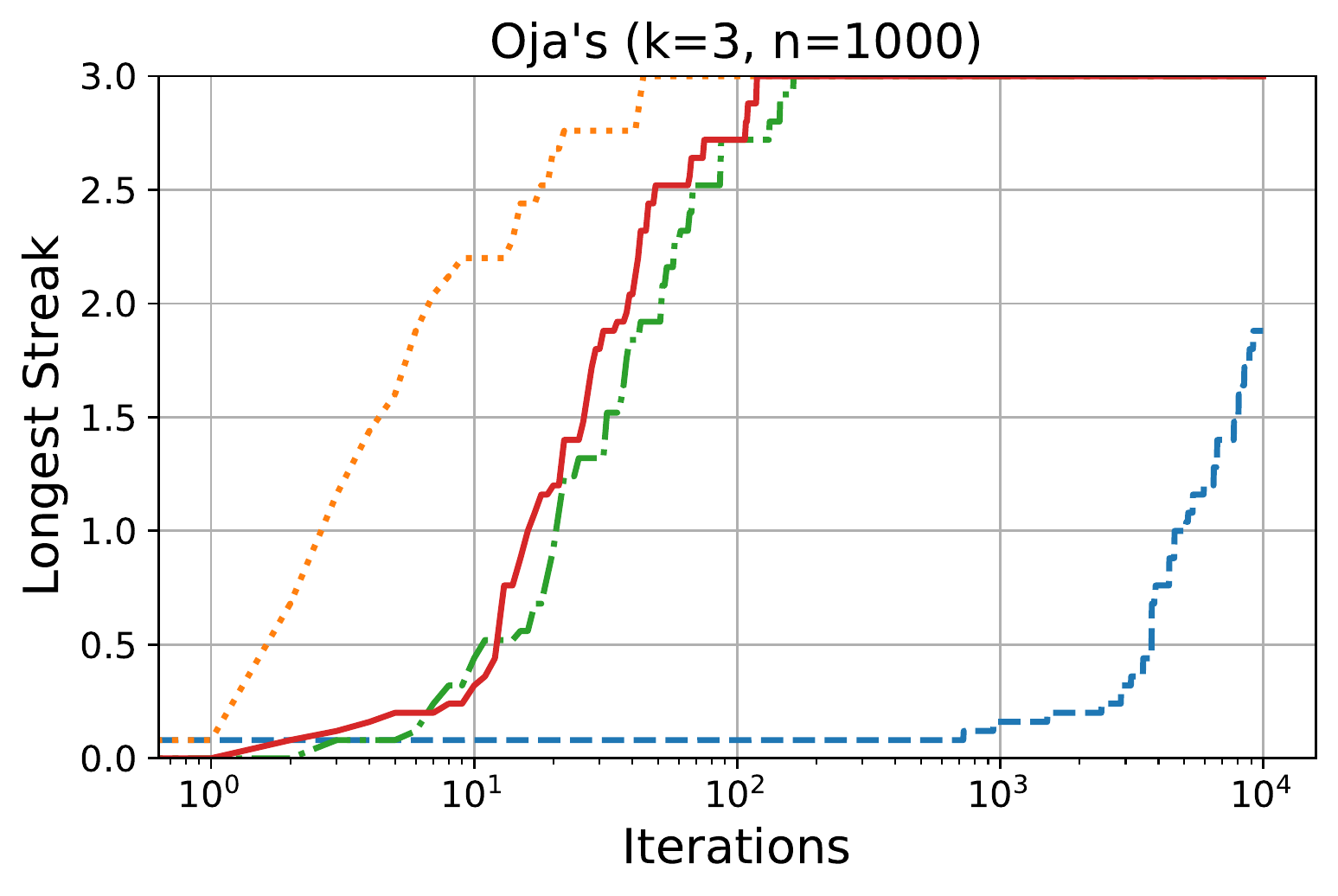}
    \includegraphics[width=0.225\textwidth]{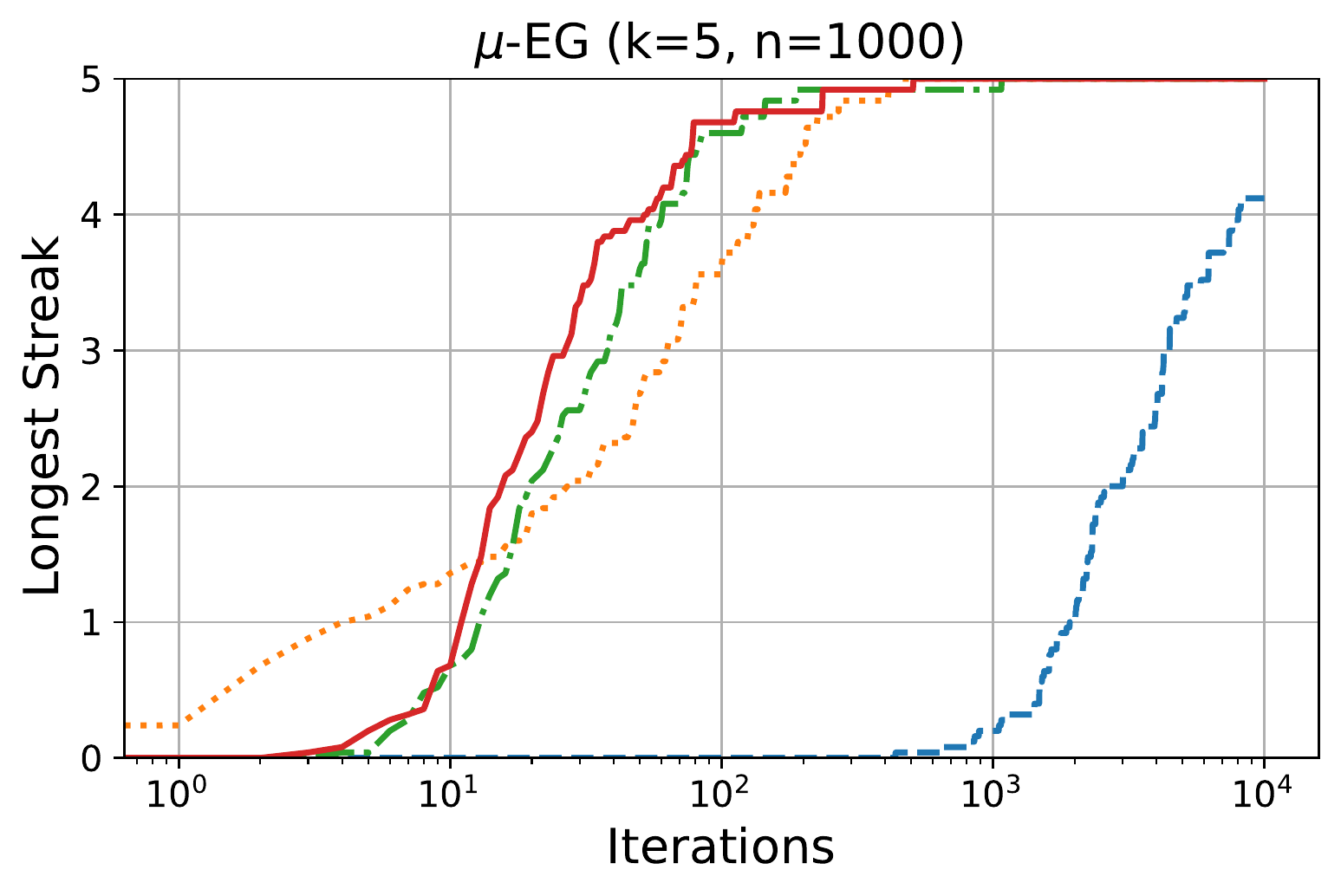}
    \includegraphics[width=0.225\textwidth]{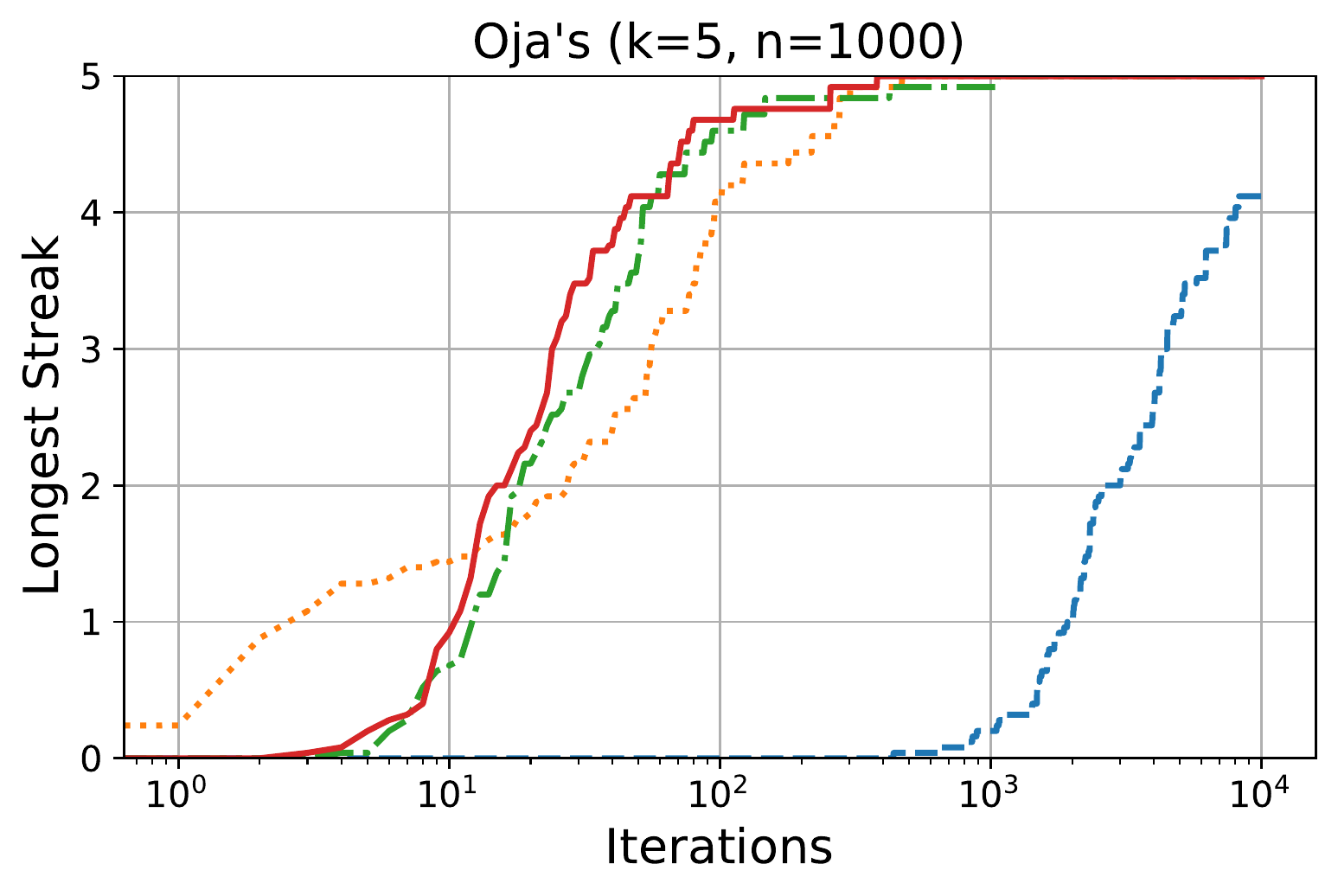}
    \includegraphics[width=0.225\textwidth]{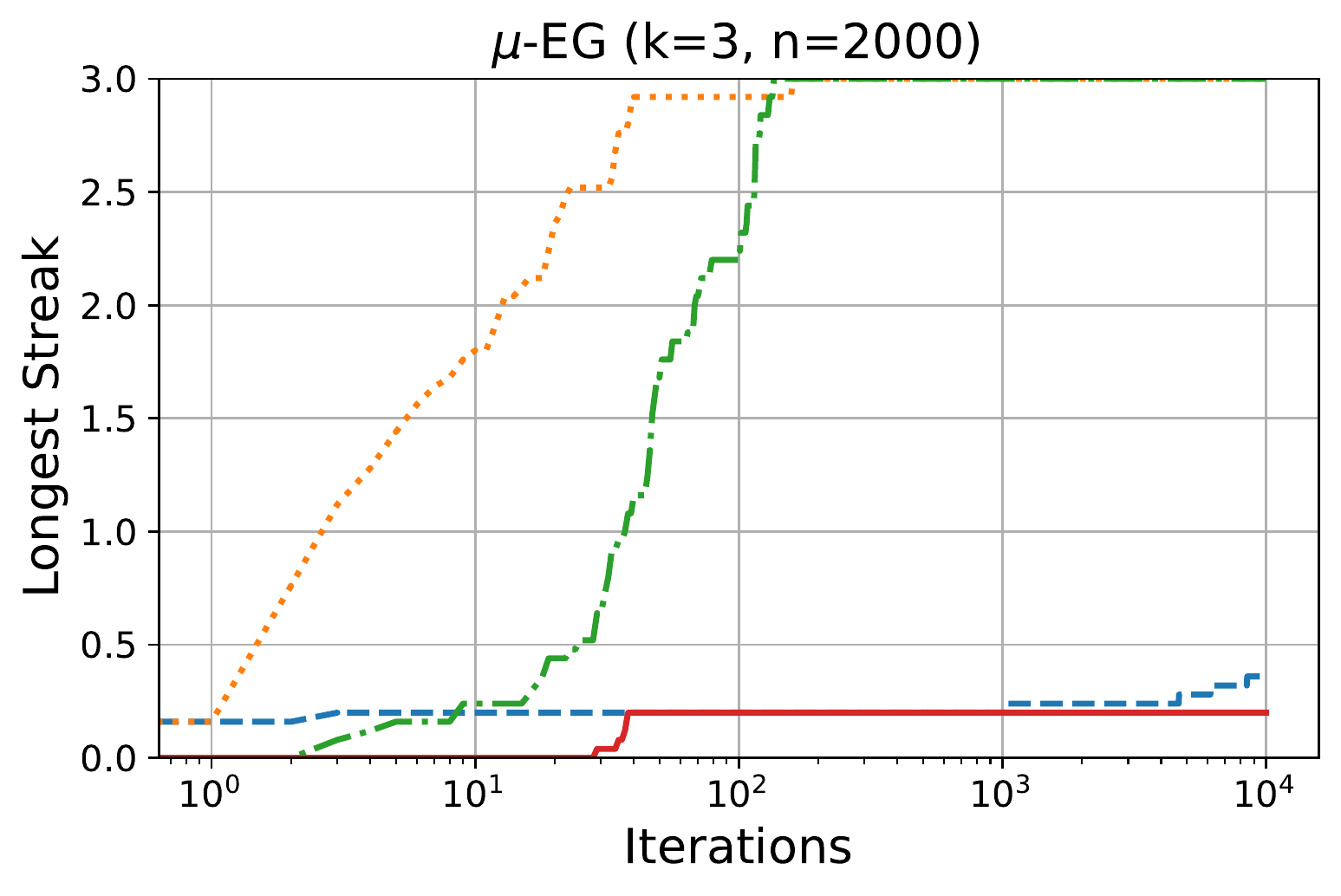}
    \includegraphics[width=0.225\textwidth]{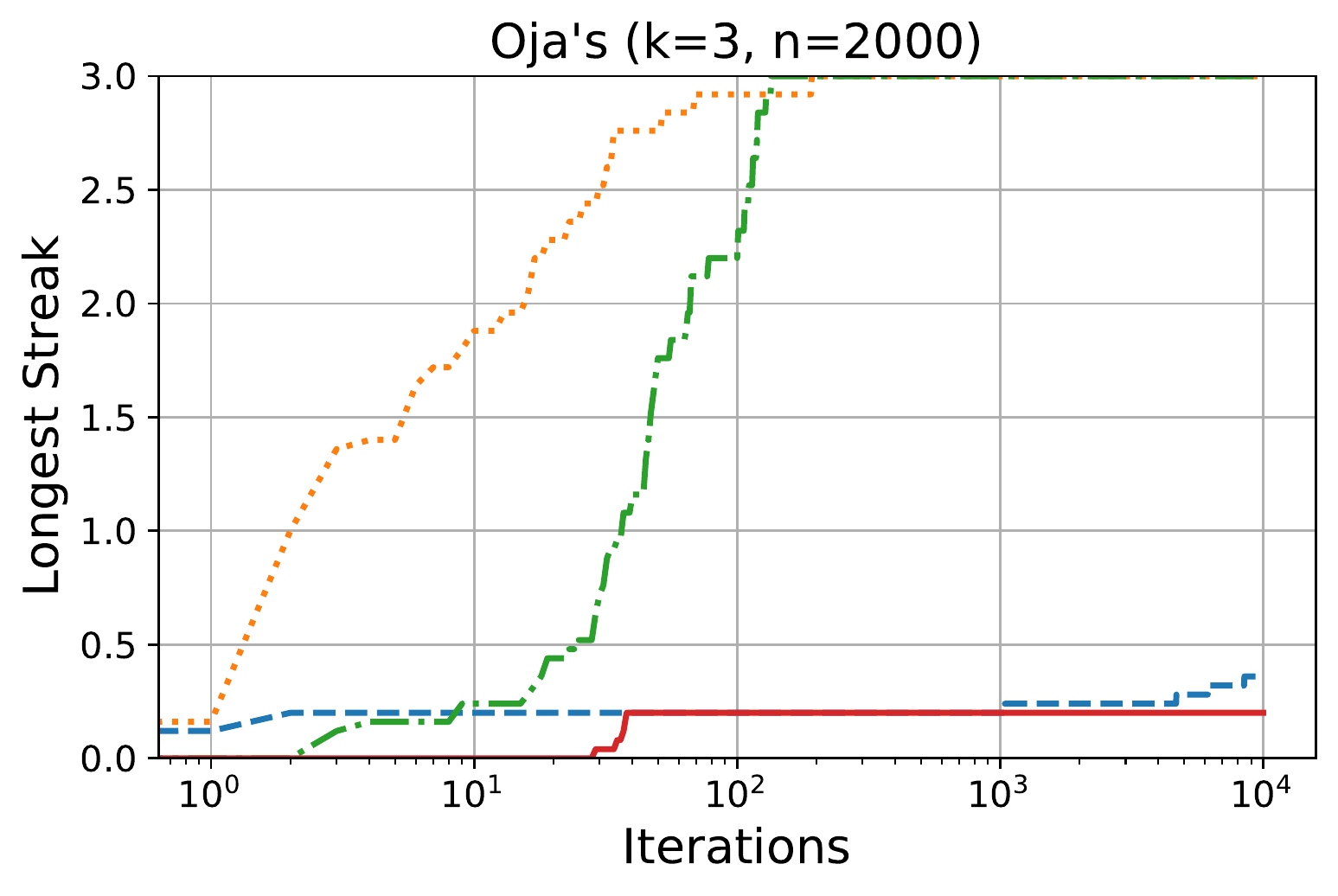}
    \includegraphics[width=0.225\textwidth]{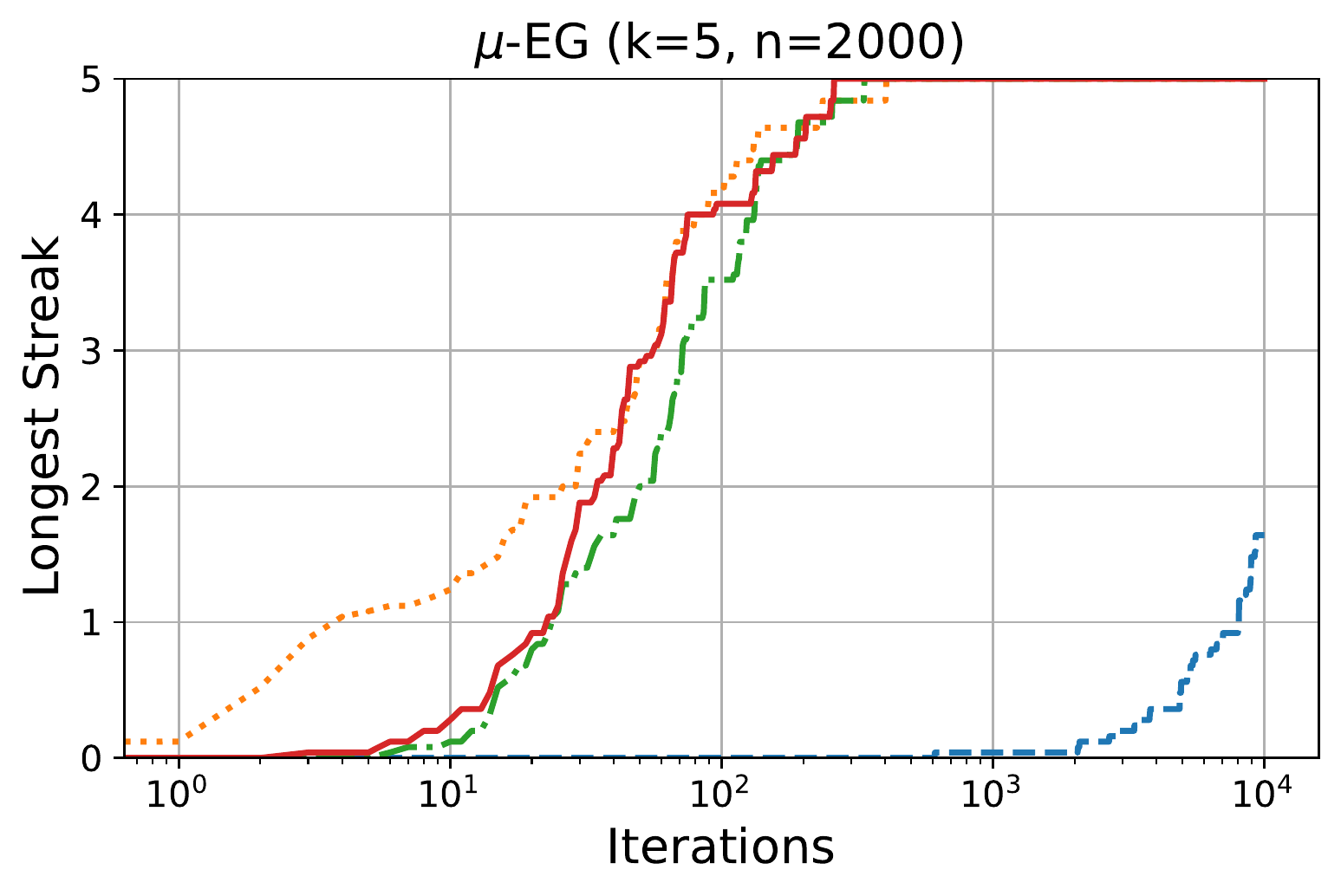}
    \includegraphics[width=0.225\textwidth]{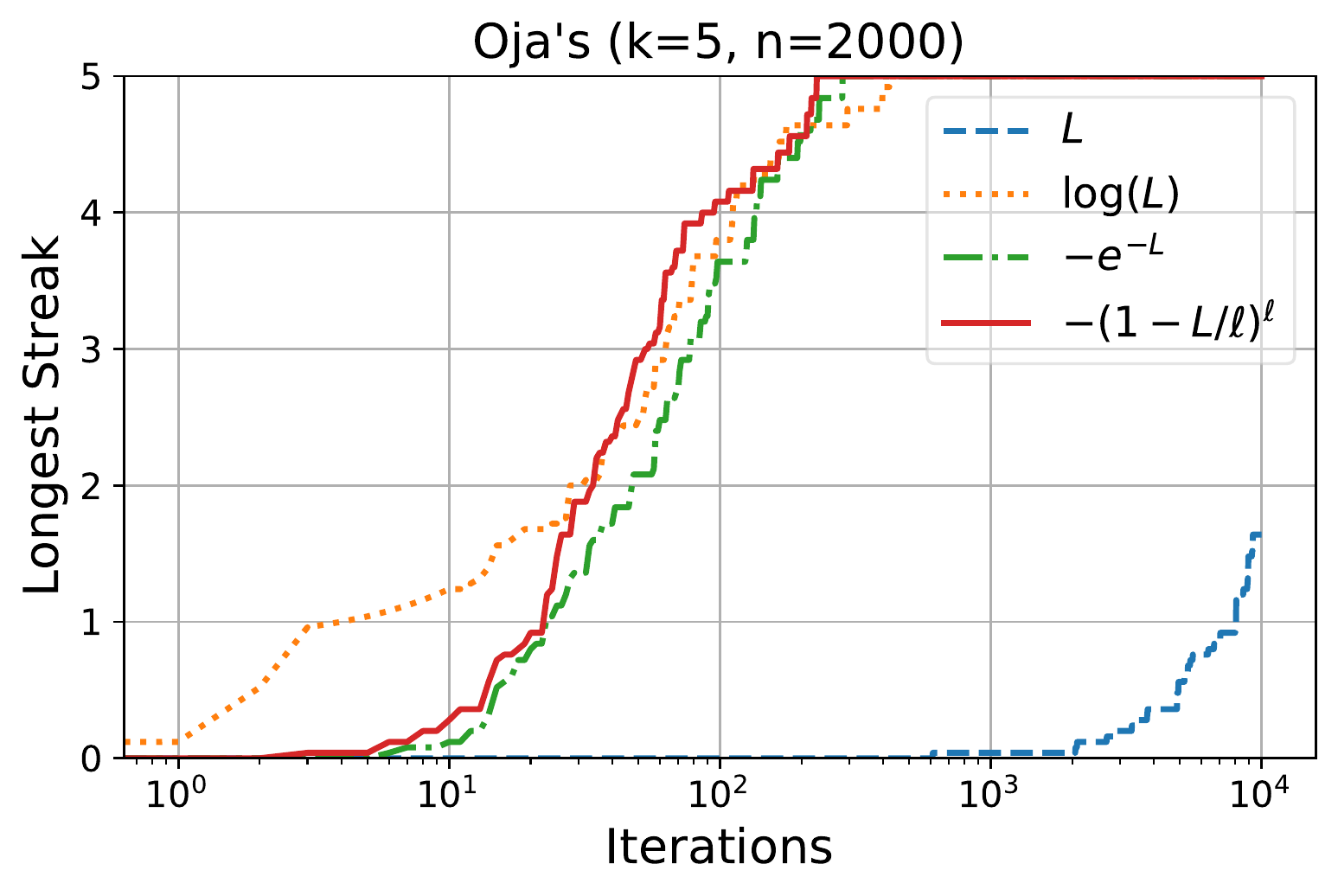}
    \caption{Cliques. Longest eigenvector streak (higher is better) is plotted over training for two different scalable SVD methods: $\mu$-EG and Oja's algorithm. Three nonlinear matrix transformations of matrix $L$ are compared against the identity transformation. One transformation (in red) is a series approximation of degree $\ell=251$ to the exact operation (in green).}
    \label{fig:cliques}
\end{figure}

Figure~\ref{fig:cliques} reveals qualitatively similar results to Figure~\ref{fig:mdp}. The series approximation works in all cases in graphs with 1000 nodes and 5 or fewer clusters. However, we can see two settings where the series approximation to the exponential transformation fails to perform well. In both settings, the number of nodes in the graph is 2000 and the number of clusters is less than 5. With 5 clusters, the series approximation succeeds. Fewer clusters while keeping the total number of nodes constant implies the individual clusters have a much higher number of edges. In other words, the maximum degree in the graph blows up. The spectral radius, maximum eigenvalue of the graph Laplacian, is upper bounded by two times the max degree. Hence, with a higher max degree we expect a larger spectral radius. In order for a series approximation to perform similarly to its exact counterpart, the approximation must contain enough terms in its polynomial to be accurate over the range $[0, 2 \texttt{deg}^*]$. Therefore, our hypothesis is that the series approximation requires more terms to be accurate enough to accelerate convergence. We perform an additional experiment in Section~\ref{sec:vary_approx} that supports this hypothesis.

\section{Conclusion}\label{sec:conclusion}
In this work, we propose a general approach to dilating eigengaps of symmetric matrices ($X^\top X$) to accelerate convergence under a stochastic optimization model (samples of $X$ are fed to the algorithm in minibatches). We discuss this approach in the context of eigendecomposition of the graph Laplacian, the SVD of a sparse matrix with a specific structure, and show how to parallelize polynomials that approximately dilate the spectrum of the matrix. We also show how this can be accomplished with samples in an unbiased way. We apply this approach to spectral-clustering and other related settings such as approximating proto-value functions and clustering a graph completed with link prediction.

In future work, we are interested in \emph{meta-learning} the coefficients of the polynomials to accelerate convergence and improving upon the simple rejection sampling scheme for generating unbiased random walks. We also aim to run stochastic, parallelized experiments at scale on much larger graphs and explore variance reduction techniques.



\section*{Acknowledgements}
We thank Daniel A. Spielman and Daniele Calandriello for sharing their expertise and pointing us to related work.

\bibliography{bib}
\bibliographystyle{icml2022}

\appendix
\section{Additional Experiments}

\subsection{Clustering Probabilistic Graphs}
\label{sec:cliques_prob}

The edges in graphs can encode relationships between entities, e.g, authors, concepts, random variables. Link prediction aims to predict missing relationships between these entities given knowledge of the known edges. Often these predictions are probabilistic, which means the completed graph contains non-binary edge weights $w_e \in [0, 1]$. We test here whether spectral clustering run with our proposed series expansion can uncover the clusters of a weighted graph given by the completion of a probabilistic link prediction engine. 

To explore the performance of our approach in this setting, we use \emph{common neighbors link prediction}, a simple and popular technique~\citep{martinez2016survey}. We generate graphs as before in Section~\ref{sec:cliques}. We then remove edges from the graph with probability $p=0.2$. Finally, we predict scores for the removed edges using \emph{common neighbors} and then normalize the scores over all missing edges to produce probabilities.



\begin{figure}
    \centering
    \includegraphics[width=0.225\textwidth]{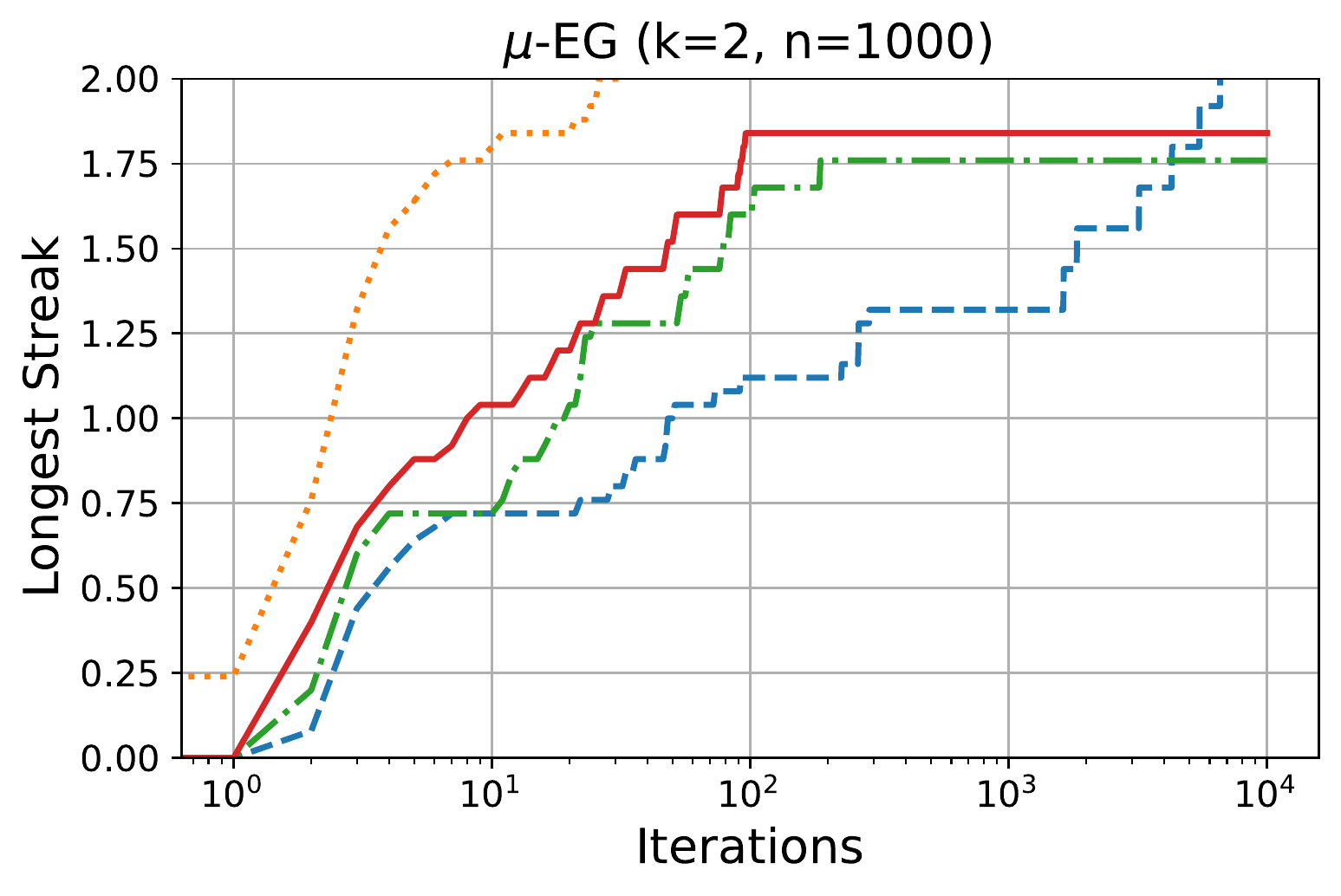}
    \includegraphics[width=0.225\textwidth]{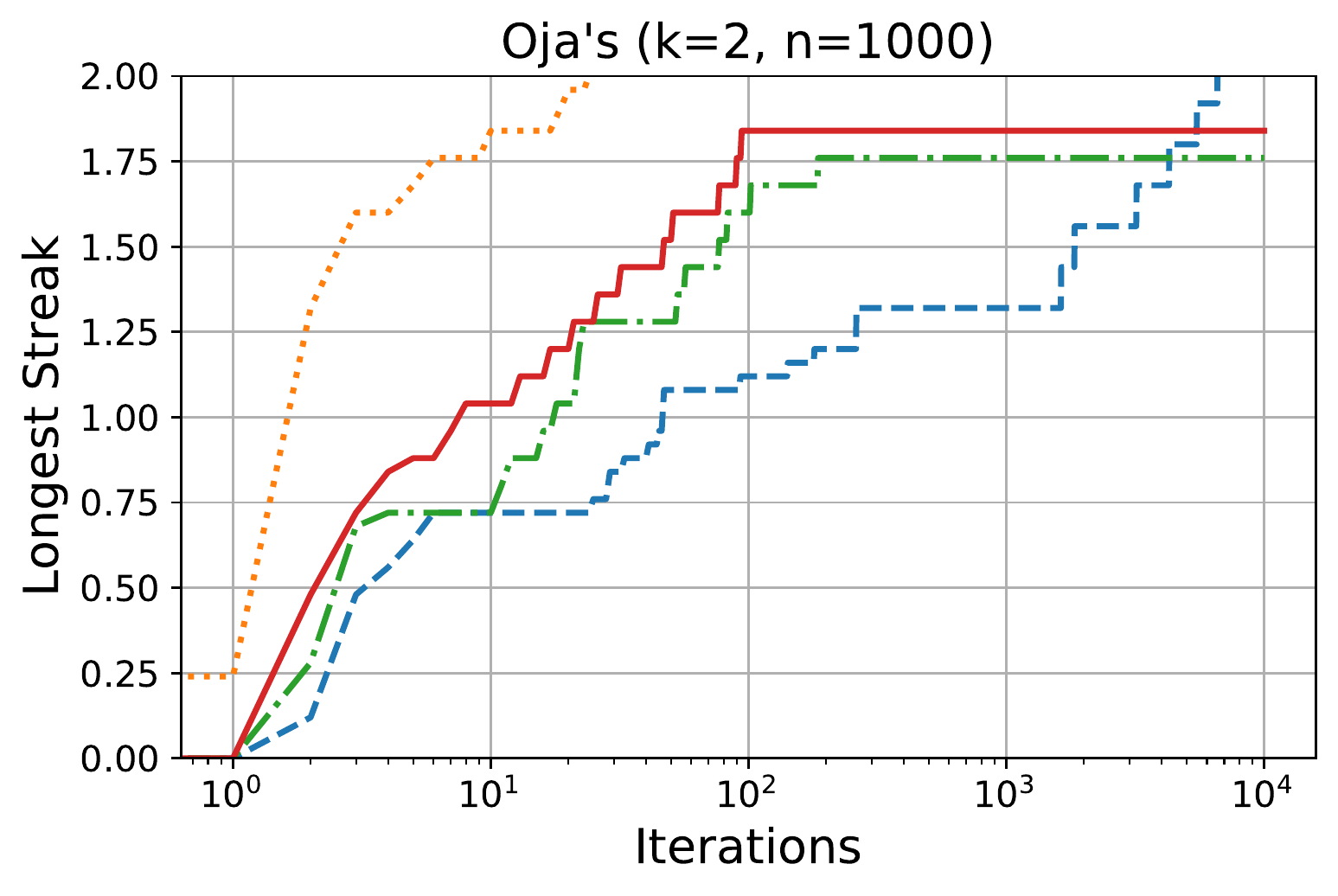}
    \includegraphics[width=0.225\textwidth]{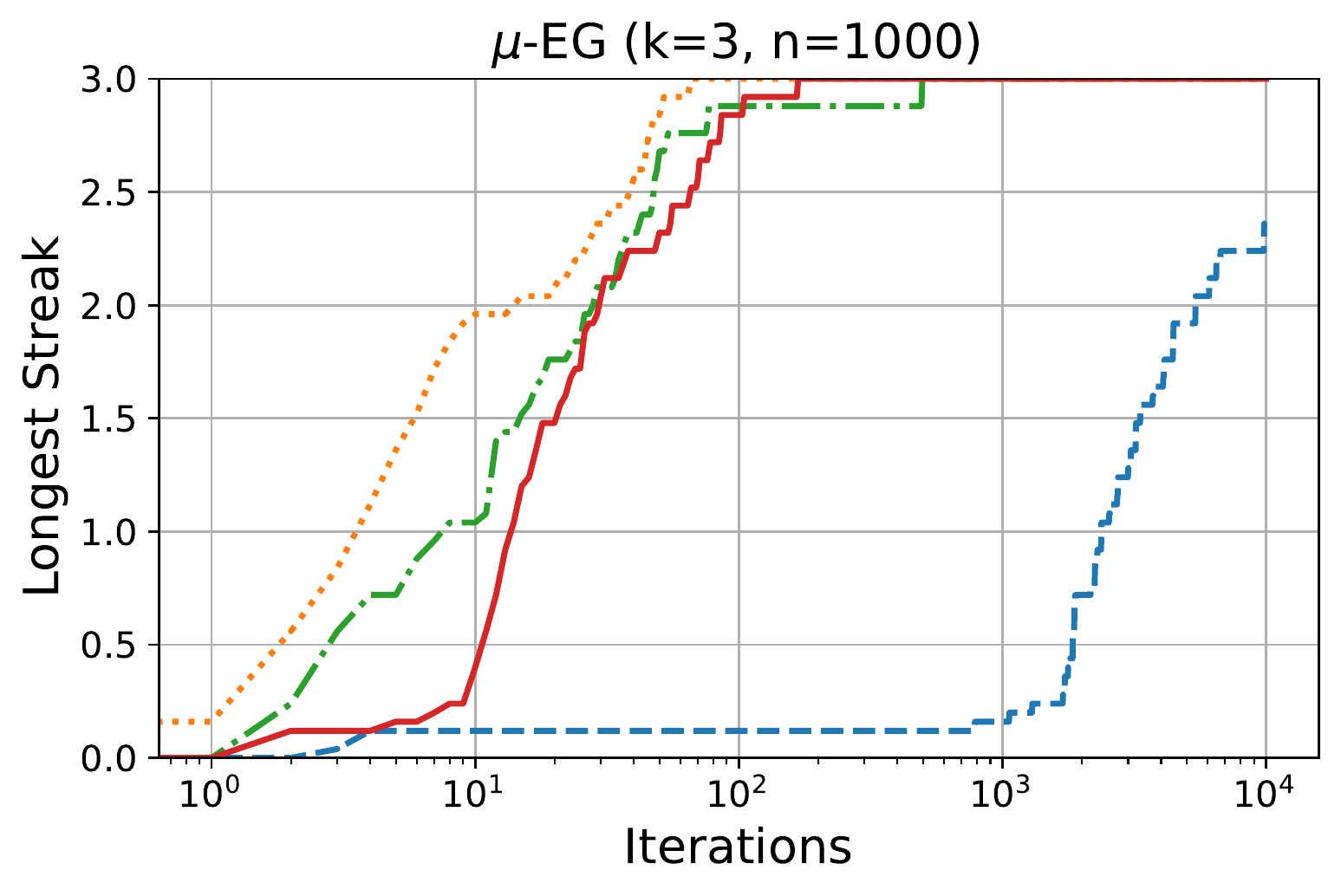}
    \includegraphics[width=0.225\textwidth]{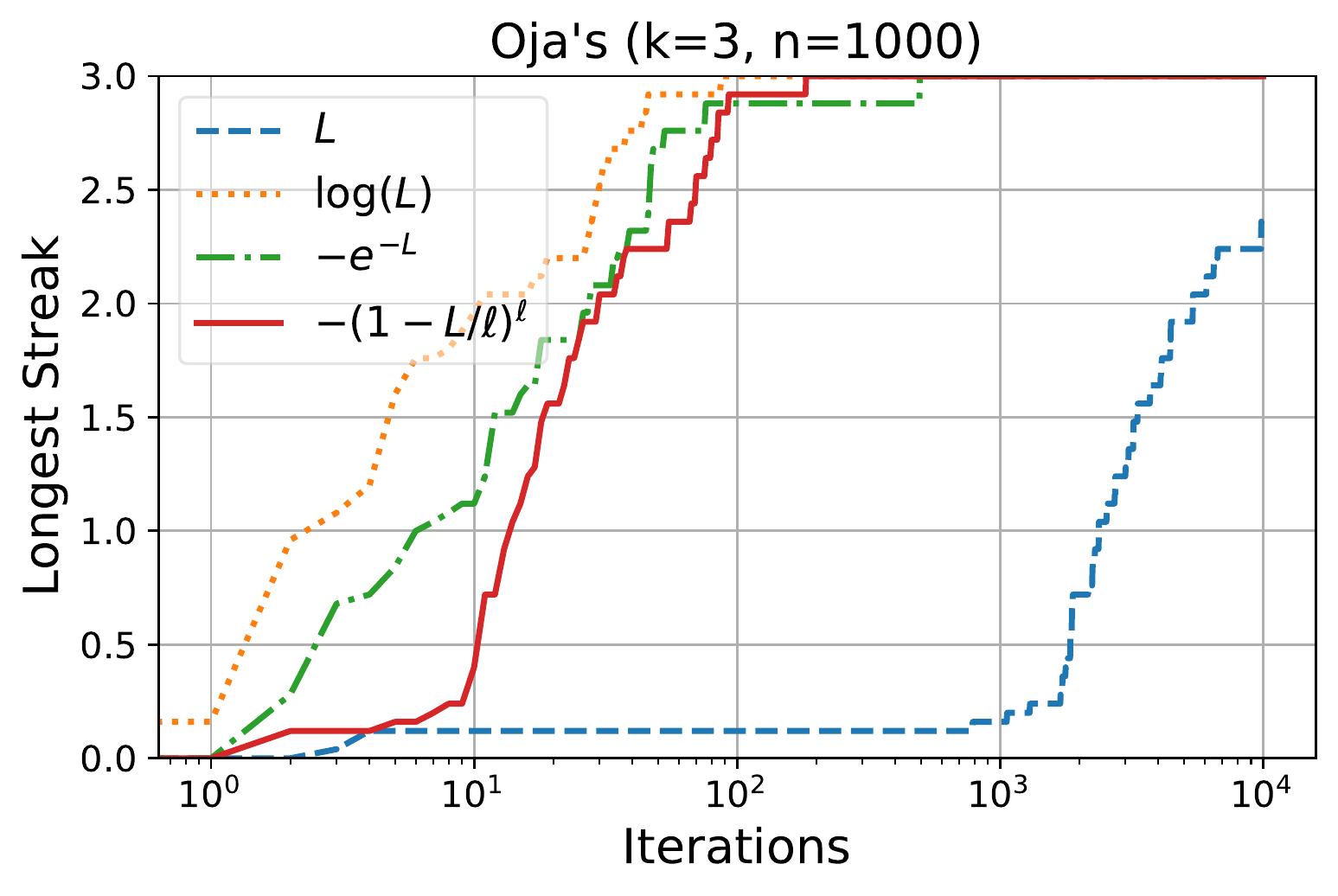}
    \caption{Link Prediction. Longest eigenvector streak (higher is better) is plotted over training for two different scalable SVD methods: $\mu$-EG and Oja's algorithm. Three nonlinear matrix transformations are compared against the identity transformation. One transformation (in red) is a series approximation of degree $\ell=251$ to the exact operation (in green).}
    \label{fig:link_prediction}
\end{figure}

Figure~\ref{fig:link_prediction} shows that \ack{} generalizes to weighted undirected graphs. This is expected as the series approximation is concerned primarily with the spectrum of the graph Laplacian and not directly with the underlying graph object.

\subsection{Accuracy of Series Approximation}
\label{sec:vary_approx}

\begin{figure}
    \centering
    \includegraphics[width=0.225\textwidth]{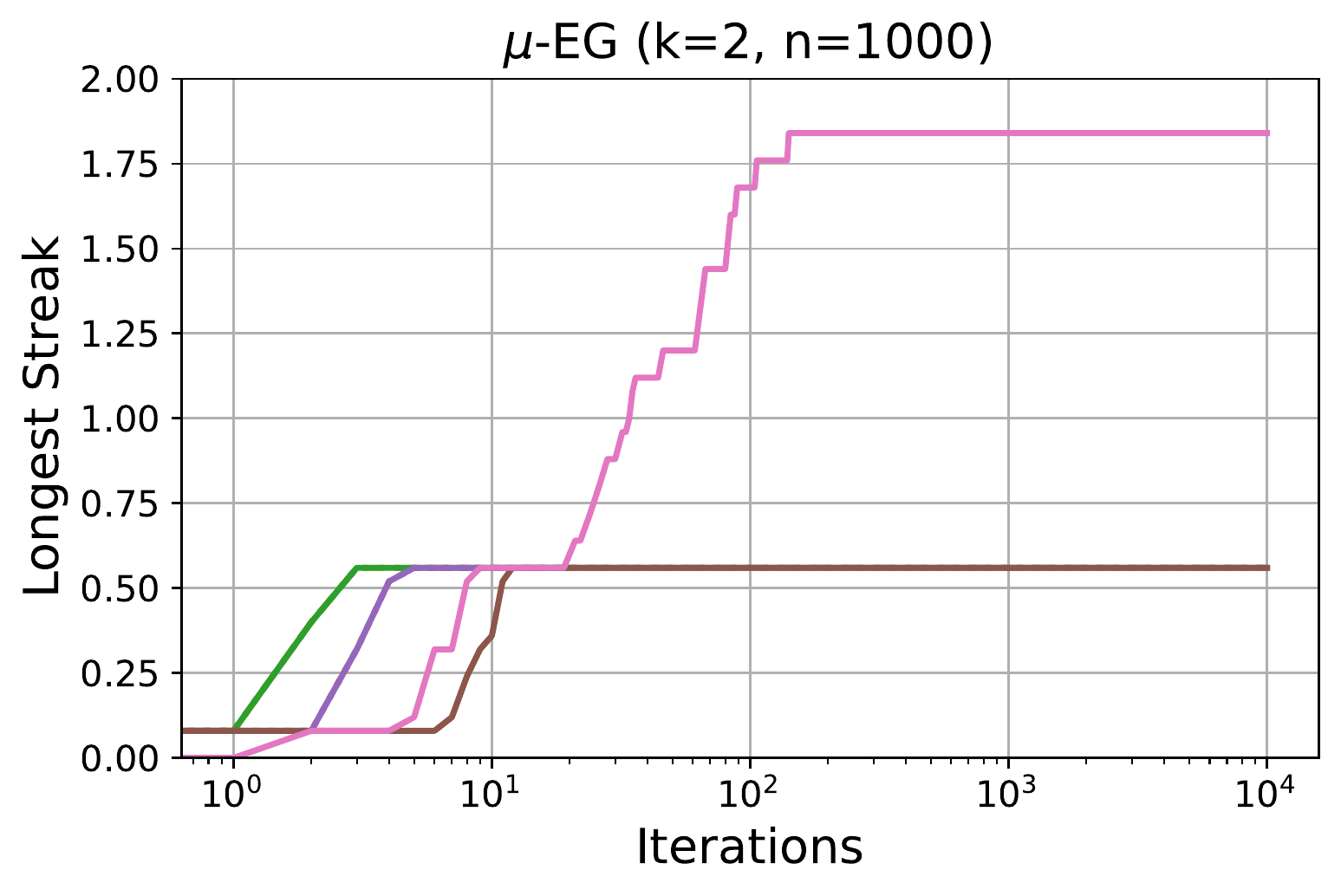}
    \includegraphics[width=0.225\textwidth]{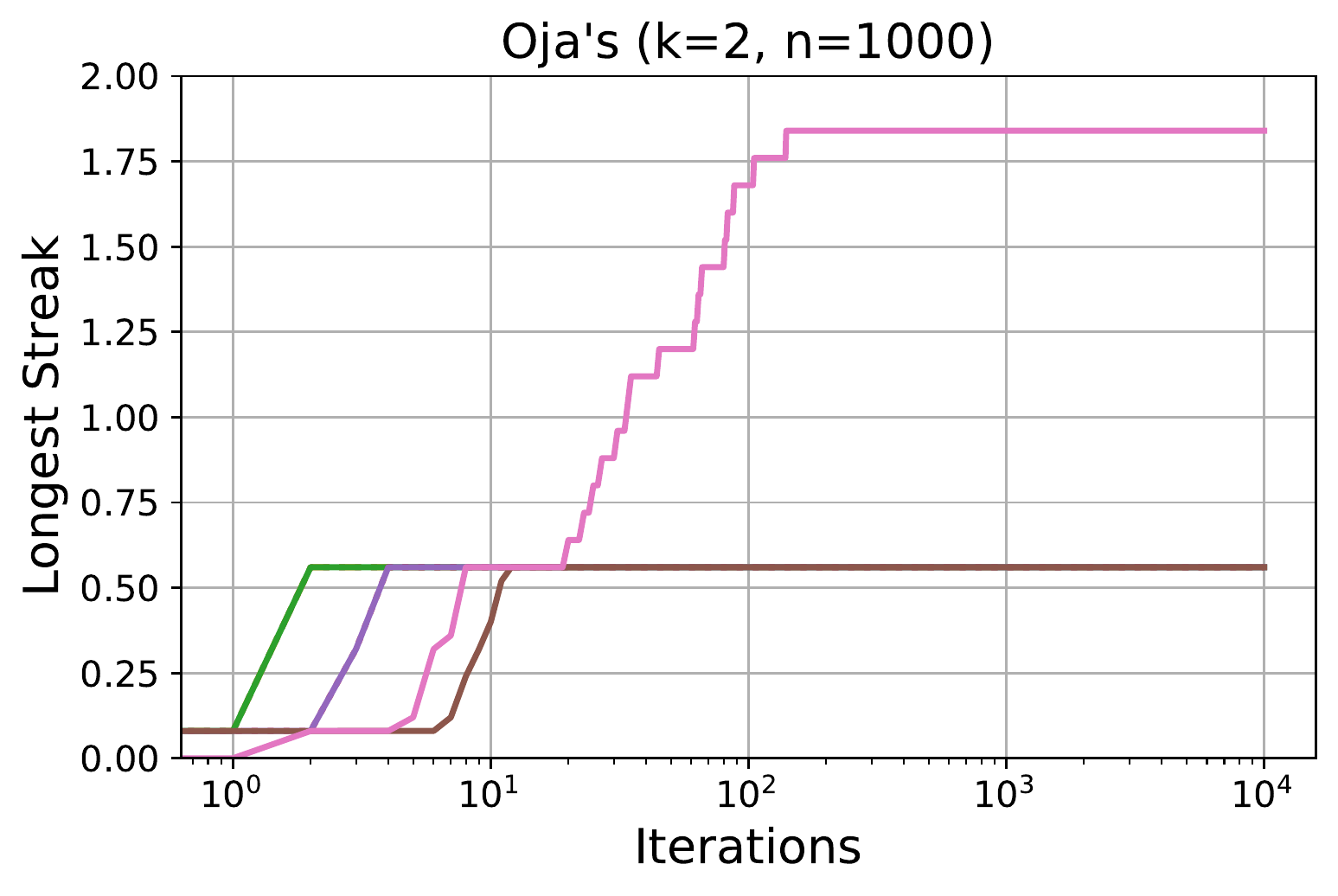}
    \includegraphics[width=0.225\textwidth]{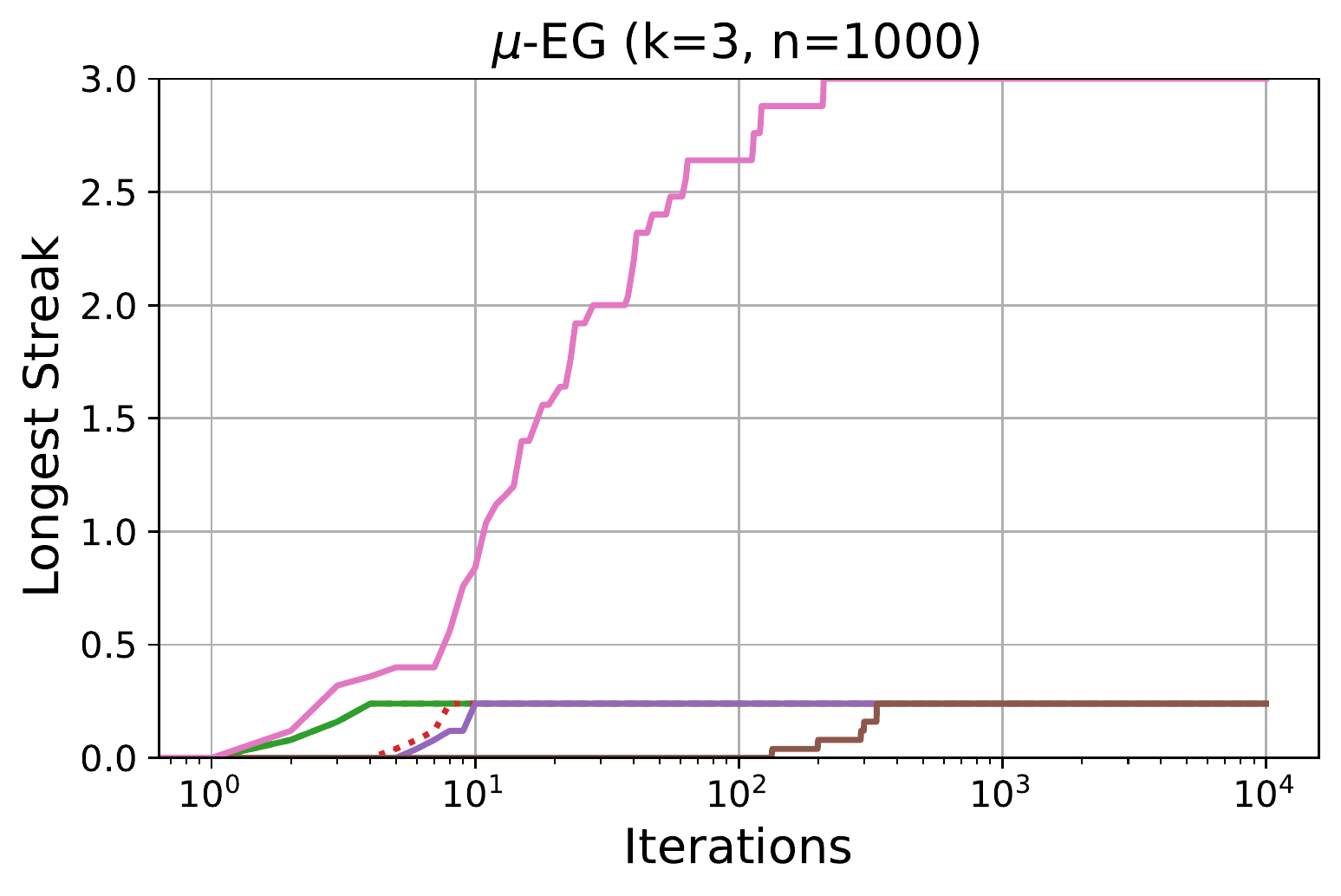}
    \includegraphics[width=0.225\textwidth]{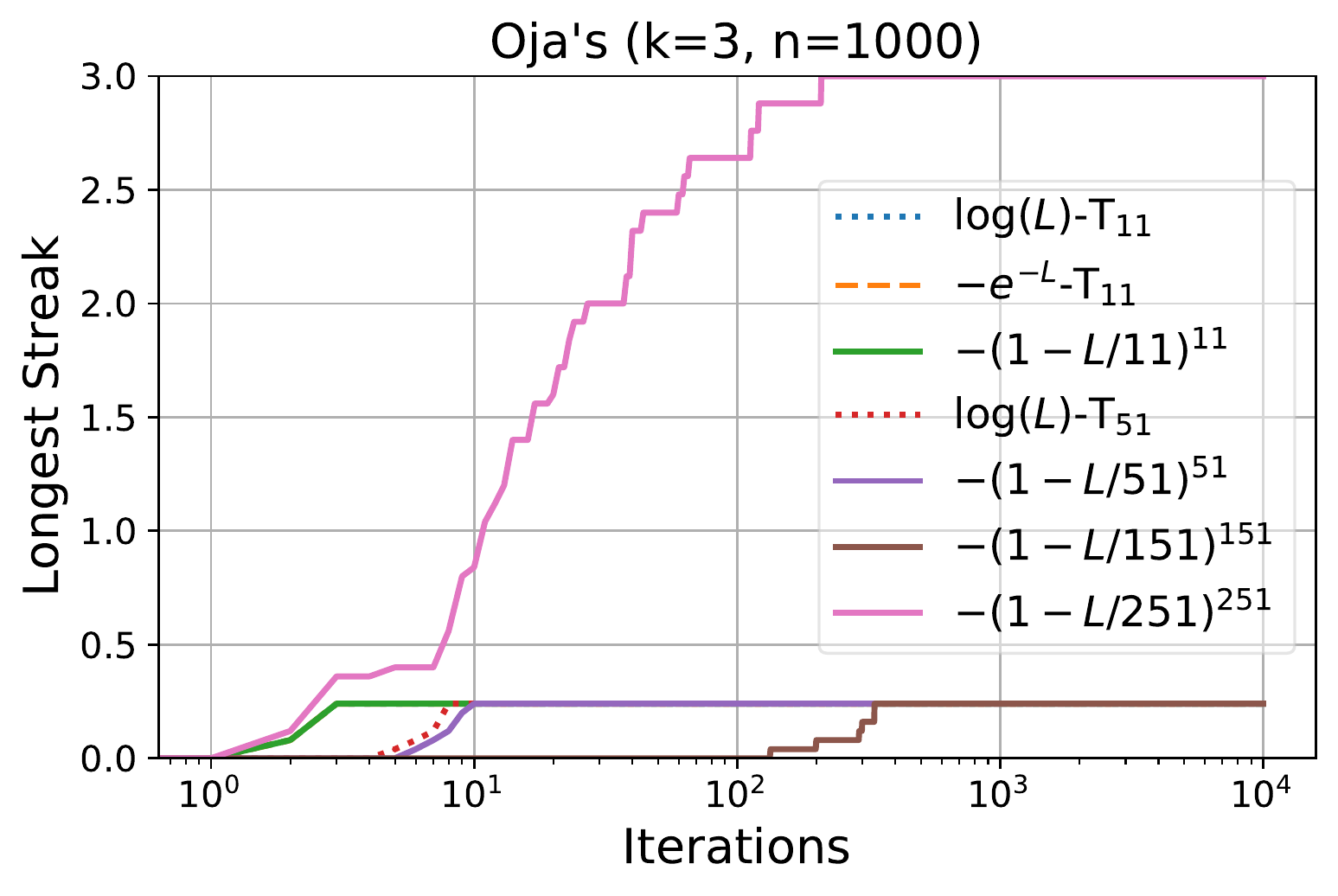}
    \caption{Cliques with Varying Series Approximation. Longest eigenvector streak (higher is better) is plotted over training for two different scalable SVD methods: $\mu$-EG and Oja's algorithm. Three types of series approximations to nonlinear matrix transformations are compared. A $T_\ell$ suffix indicates an $\ell$-degree Taylor series expansion was used.}
    \label{fig:cliques_approx}
\end{figure}

In this experiment we observe the results of varying the number of terms in the series approximation. We consider the same graph generation process as in Section~\ref{sec:cliques}.

Figure~\ref{fig:cliques_approx} reveals two conclusions. One is that if the number of terms in the series approximation is insufficient, it will fail to accelerate convergence. For example, the limit series approximation of the exponential fails for $\ell=11$, $51$, and $151$ terms but succeeds for $251$. The other is that the limit series approximation of the exponential outperforms the other series approximations we explore. This result was true across experiments.

\section{Related Work}\label{sec:related}

Eigenvectors of the graph Laplacian are not just important for spectral clustering. There has been intense interest in computing the top eigenvector, also known as the Perron vector, as it represents the stationary distribution of a Markov chain on the graph. This is the core challenge behind the famous PageRank algorithm~\citep{page1999pagerank}. Eigenvectors of the graph Laplacian are also useful as representations for reinforcement learning (RL). However, other works focus more on alternative bases or representations than on accelerating convergence to the eigenvectors of the graph Laplacian~\citep{mahadevan2005proto, petrik2007analysis, machado2017laplacian}. Special attention has been paid to solving linear equations with the graph Laplacian as well, $Lx = b$ where $b$ is a given vector. The graph Laplacian has a special structure that falls into the more general class of $M$-matrices. \citet{ahmadinejad2019perron} developed nearly linear time solutions to solving the above linear equation and finding the Perron vector.

Other approaches to finding the Perron vector include \emph{shift-and-invert} preconditioning~\citep{garber2016faster} where rather than solving for the Perron vector of $L$, we instead search for the Perron vector of $(\lambda^* - A)^{-1}$. The motivation is the same as ours in this work \textemdash dilate the eigengap. Let the top two eigenvalues of $L$ be $\lambda_{\vert \mathcal{V} \vert}$ and $\lambda_{\vert \mathcal{V} \vert - 1}$ with eigengap $g_n = \lambda_{\vert \mathcal{V} \vert} - \lambda_{\vert \mathcal{V} \vert - 1} = \bar{g}_n \lambda_{\vert \mathcal{V} \vert}$. Also set $\lambda^* = \lambda_{\vert \mathcal{V} \vert} + g_n$. Then the number of steps required for convergence for iterative stochastic approaches\footnote{See Table 1 of~\citep{allen2017first}.} on the original $L$ is generally $\mathcal{O}(1 / \bar{g}^p)$ for $p \ge 1$ while the steps required for the new shifted and inverted Laplacian is $\mathcal{O}(1)$. The inversion step is solved approximately with least squares. However, these results do not solve the problem of finding the bottom-$k$ eigenvectors with $k > 1$.

Other works aim to learn a specific subspace of the graph Laplacian rather than the actual eigenvectors. \citet{frostig2016principal} approximates the matrix step function with a polynomial series to accelerate principal component regression. Likewise, \citet{tremblay2016compressive} uses a polynomial approximation to a matrix step function to accomplish ``fast graph filtering'' for accelerating spectral clustering. \citet{di2016efficient} uses Chebyshev polynomials to approximate a similar interval indicator function to count the number of eigenvalues in a given range. This can be used to split eigendecomposition into computing distinct subspaces separately. \citet{shuman2011chebyshev} also approximates graph operators with Chebyshev polynomials for distributed computation in wireless sensor networks. 

These works apply to spectral clustering, but do not consider the typical stochastic optimization setting where graph data (edges) are fed in minibatches to an iterative solver. In the case where the graph is too large to fit in memory or the graph is a random variable (edges are sampled from a distribution, e.g., the output of a link prediction model), these methods are not well-suited. \citet{saade2014spectral} proposes a linear function of the diagonal degree matrix, $D$ and adjacency matrix, $A$, called the \emph{Bethe} Hessian for spectral clustering of graphs generated by the stochastic block model (SBM)~\citep{holland1983stochastic}. In \citep{cohen2014preconditioning}, a matrix preconditioner is constructed in expectation via samples. This preconditioner is then inverted and used to solve a linear system by modified Richardson iteration method. Distributed methods for PCA have been developed, but these focus on finding the top-$k$ eigenvectors (or rather, top-$k$ subspace) and are not tailored to the spectra of graph Laplacians~\citep{raja2020distributed,gang2019fast} nor do they focus on dilating eigengaps.

Lastly, graph sparsification uses the graph Laplacian of a similar, but sparser graph constructed by e.g., removing edges to accelerate eigenvector computation. An optimal sparsifier can be found in near linear time but requires a pass through the entire dataset. For every weighted graph $G = (\mathcal{V}, \mathcal{E}, w)$ and every $\epsilon > 0$, there is a re-weighted subgraph of $G$ with $\tilde{\mathcal{O}}(\vert \mathcal{V} \vert / \epsilon^2)$ edges that is a $(1+\epsilon)$ approximation of $G$ where the $\tilde{\mathcal{O}}$ notation hides log factors. It is possible to find such a subgraph in $\tilde{\mathcal{O}}(\vert \mathcal{E} \vert)$ time~\citep{spielman2011spectral}. \citet{jindal2015efficient} parallelize graph sparsification but on a fixed dataset. Note that this line of work is orthogonal to the approach we take here. A nonlinear transformation can still be applied to the sparse graph to accelerate convergence even further.

While components of the approach we take here are known \textemdash general eigengap dilation (e.g., shift-and-invert) and the connection between powers of the graph Laplacian and random walks \textemdash our approach combines these in a way to tackle approximating the bottom-$k$ eigenvectors of the graph Laplacian at scale under the stochastic optimization model (or SBM). We discuss the key challenges that arise in this setting, namely unbiased estimates of random walks and parallelization. Also, unlike previous works we empirically evaluate our approach under a variety of transformation functions, to our knowledge not previously considered, and identify a promising candidate in the decaying exponential.

\end{document}